\documentclass[lettersize,journal]{IEEEtran}
\usepackage[utf8]{inputenc}

\usepackage{amsmath}
\usepackage{amssymb}
\usepackage{algorithmic}
\usepackage{algorithm}
\usepackage{array}
\usepackage[caption=false,font=normalsize,labelfont=sf,textfont=sf]{subfig}
\usepackage{textcomp}
\usepackage{stfloats}
\usepackage{url}
\usepackage{verbatim}
\usepackage{grffile}
\usepackage{graphicx}
\usepackage{cite}
\usepackage{hyperref} 
\usepackage{booktabs}
\usepackage{tabularx}
\usepackage{tablefootnote}
\usepackage{orcidlink}

\hypersetup{
    colorlinks=true,
    linkcolor=blue,
    citecolor=blue,
    filecolor=magenta,
    urlcolor=cyan
}

\begin{document}




\title{Transparent Fragments Contour Estimation via Visual-Tactile Fusion for Autonomous Reassembly}
\author{Qihao Lin\,\orcidlink{0009-0001-3055-6208}, Borui Chen\,\orcidlink{0009-0004-6613-4033}, Yuping Zhou\,\orcidlink{0009-0004-2535-9287}, Jianing Wu\,\orcidlink{0000-0003-0902-4466}, Yulan Guo\,\orcidlink{0000-0001-7051-841X}, Weishi Zheng\,\orcidlink{0000-0001-8327-0003}, and Chongkun Xia\,\orcidlink{0000-0001-5396-7643}
\thanks{This work was supported by the National Key R\&D Program of China (2024YFB3213600),  National Natural Science Foundation of China (No. 62203260), Guangdong Basic and Applied Basic Research Foundation (2023A1515011773).}
\thanks{Qihao Lin, Borui Chen, Yuping Zhou, Jianing Wu, and Chongkun Xia are with the School of Advanced Manufacturing, Sun Yat-sen University, Shenzhen 518107, China.}
\thanks{Yulan Guo is with School of Electronics and Communication Engineering, Sun Yat-sen University, Shenzhen 518107, China.}
\thanks{Weishi Zheng is with the School of Computer Science and Engineering, Sun Yat-sen University, Guangzhou 510006, China.}
\thanks{Corresponding author: Chongkun Xia (xiachk5@mail.sysu.edu.cn).}}




\maketitle

\begin{abstract}
The contour estimation of transparent fragments is very important for autonomous reassembly, especially in the fields of precision optical instrument repair, cultural relic restoration, and identification of other precious device broken accidents. Different from general intact transparent objects, the contour estimation of transparent fragments face greater challenges due to strict optical properties, irregular shapes and edges. To address this issue, a general transparent fragments contour estimation framework based on visual-tactile fusion is proposed in this paper. First, we construct the transparent fragment dataset named TransFrag27K, which includes a multiscene synthetic data of broken fragments from multiple types of transparent objects, and a scalable synthetic data generation pipeline. Secondly, we propose a visual grasping position detection network named TransFragNet to identify, locate and segment the sampling grasping position. And, we use a two-finger gripper with Gelsight Mini sensors to obtain reconstructed tactile information of the lateral edge of the fragments. 
By fusing this tactile information with visual cues, a visual–tactile fusion material classifier is proposed. Inspired by the way humans estimate a fragment's contour combining vision and touch, we introduce a general transparent fragment contour estimation framework based on visual-tactile fusion, 
demonstrates strong performance in real-world validation. Finally, a multi-dimensional similarity metrics based contour matching and reassembly algorithm is proposed, providing a reproducible benchmark for evaluating visual–tactile contour estimation and fragment reassembly. The experimental results demonstrate the validity of the proposed framework. The dataset and codes are available at https://github.com/Keithllin/Transparent-Fragments-Contour-Estimation.
\end{abstract}

\begin{IEEEkeywords}
Transparent fragment, reassembly, contour estimation, visual-tactile fusion
\end{IEEEkeywords}
\section{Introduction}
\IEEEPARstart{T}{ransparent} fragments, produced by the fracture of glass, resin, or plastic, are widely encountered but easily overlooked in real-world scenarios. Unlike intact transparent objects (such as a whole glass), which have a continuous and modelable complete surface are often detected based on prior models or category templates~\cite{{Jiang2023Review}}, these fragments exhibit highly irregular shapes, jagged edges, and uneven thicknesses, leading to noisy or missing signals in both RGB and depth sensors. Since the target shape is usually inaccurate and the fragment shapes are usually irregular, it is challenging to identify the correct pose of each part and align them accordingly for autonomous reassembly. In fact, the reassembly task of transparent fragments are usually time-consuming and skill-intensive, and sometimes even impractical if the number of parts is large~\cite{Xu2024FragmentDiff}. For sensor-based robotic systems, there is still no standardized benchmark to evaluate such tasks under real-world constraints. 
Therefore, robust contour estimation of each transparent fragment is a fundamental step toward reassembly, and a general transparent fragment contour estimation framework supports precise edge alignment and robust matching of fragments, which are crucial for restoring damaged artifacts, optimizing production processes, and understanding complex geometries.

However, existing fragment reassembly pipelines and transparent object perception algorithms both face critical limitations when applied to transparent fragments, highlighting the need for a benchmark framework for precise contour estimation. Recent learning-based methods have advanced fragment reassembly by leveraging geometric reasoning~\cite{sellan2022}, graph matching~\cite{scarpellini2024}, and equivariant neural networks~\cite{wu2023}, yet they remain limited in scenarios involving transparent fragments, where refractive distortion, irregular shapes, and fractured surfaces degrade performance. Likewise, transparent object perception methods such as contextual learning~\cite{Mei2023LargeField}, and monocular depth estimation~\cite{Liang2022Monocular} focus on complete object segmentation and often fail at fragment-level continuity or edge fidelity. Though recent works explore multi-modal strategies using polarimetric or thermal cues~\cite{Shao2024polarimetric,Huo2022GlassSeg}, their deployment is hindered by complex sensing setups and poor scalability in cluttered, fragment-rich environments. These limitations underscore the need for a general contour estimation framework that combines robust visual detection with high-resolution tactile sensing to extract accurate fragment boundaries. By bridging vision and tactile signals at the contour level, such a framework enables precise edge recovery and provides the geometric basis for reliable grasp planning, alignment, and autonomous reassembly.

\begin{figure}[!t]
\centering
\includegraphics[width=3.5in]{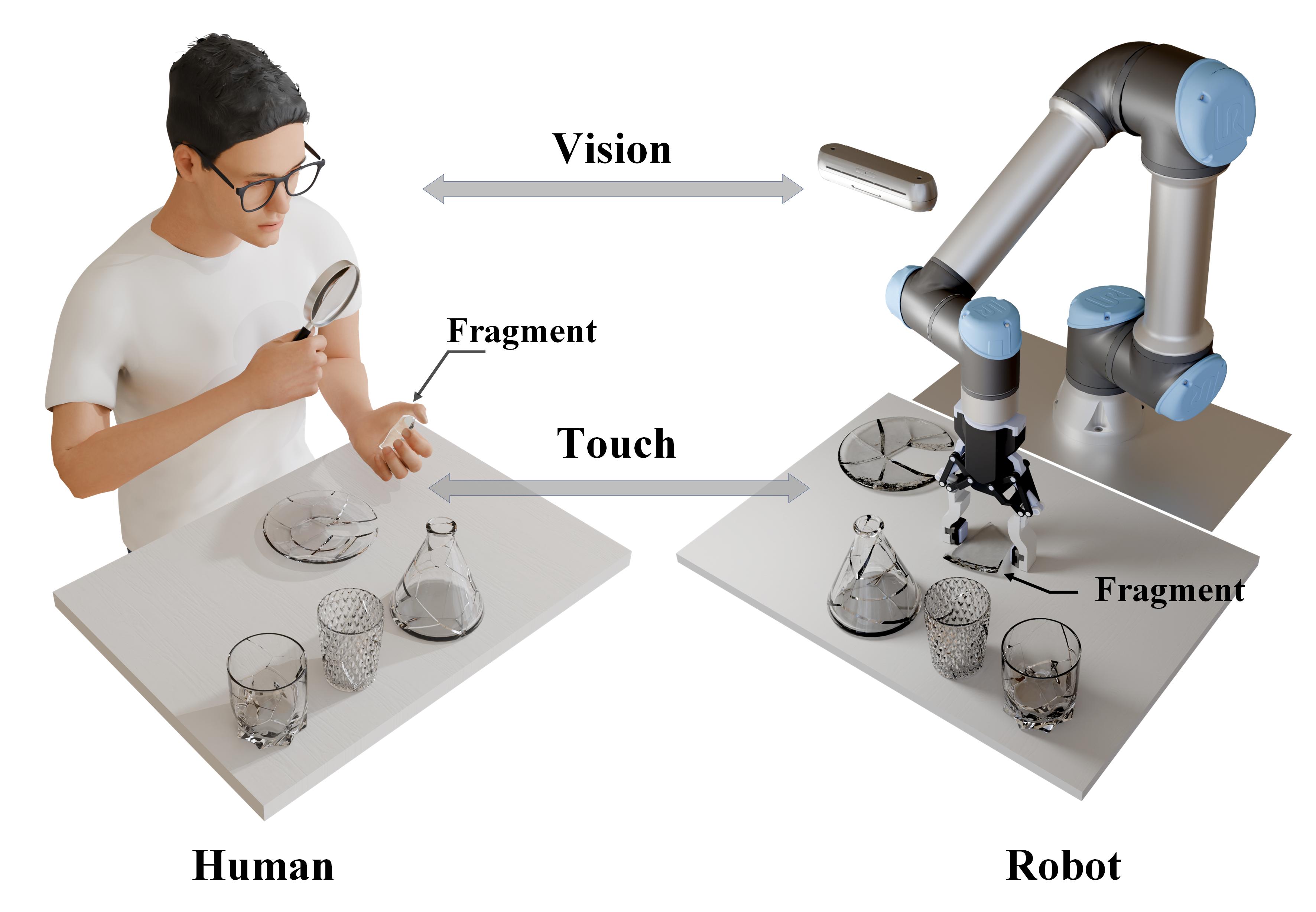}
\caption{Visual-tactile fusion contour estimation inspired by the human strategy of combining vision with touch.}
\label{human_people}
\end{figure}

Inspired by the way humans estimate fragments' contour combining vision and touch, as shown in Fig. \ref{human_people}, we introduce a general transparent fragments contour estimation framework based on visual-tactile fusion for autonomous reassembly. 
At its core, the benchmark framework integrates a GelSight Mini-equipped two-finger gripper and a detection network TransFragNet to identify contact regions and recover fine surface profiles. These tactile signals complement noisy visual cues and enable robust contour estimation through a dedicated reconstruction algorithm. Validation results show that the proposed visual mask images and tactile edge information based contour estimation achieves alignment quality of 93.8\% and 96.7\% in real-world trials. 
We further introduce a visual–tactile fusion material classification module and a contour matching and reassembly algorithm that evaluates a multi-dimensional similarity metric, including geometric complementarity, height consistency, and gradient continuity for accurate fragment alignment.
Our method demonstrates robust performance across partial and full fracture scenarios, enabling effective reassembly of transparent fragments under challenging conditions. Specifically, the contributions of this work are threefold.

\begin{itemize}
  \item We introduce a general transparent fragments contour estimation framework based on visual-tactile fusion for autonomous reassembly of transparent irregular fragments. The benchmark framework integrates a deep-learning grasp point detection network (TransFragNet), GelSight Mini-based tactile reconstruction, a visual–tactile material classifier, a contour matching and reassembly algorithm.

  \item We validated the proposed benchmark framework through several experiments, including tests on single components and overall framework. The excellent performance of the framework was demonstrated. To verify the generality and robustness of the framework, we designed a multi-material reassembly task with 20 fragments, including glass, acrylic, crystal, and plastic. The results show that our framework has considerable generalization and robustness for the fragment reassembly task.

  \item To support reproducibility and comparison, we construct TransFrag27K and a scalable synthetic data generation pipeline. TransFrag27K is the first large-scale transparent fragment dataset, comprising synthetic samples from diverse objects such as dishes, cups and beakers. The dataset includes over 27,000 annotated instances under varied lighting, backgrounds, and fragment configurations, providing a foundation for training and rigorous benchmarking.
\end{itemize}




The remainder of this paper is organized as follows.
Section II reviews related work on transparent object perception, tactile-guided manipulation, and fragment reassembly. Section III introduces the construction and annotation of TransFrag27K, our large-scale transparent fragment dataset. Section IV details the overall visual–tactile framework, covering the TransFragNet detection network, the tactile reconstruction module, the visual–tactile fusion material classification, the integrated contour estimation procedure, the contour matching and reassembly algorithm. Furthermore, experimental
validations are provided in Section V. Finally, Section VI concludes the paper and outlines future research directions.
\section{Related Work}

\subsection{Transparent Object Detection and Grasping}
With the advent of deep learning, segmentation networks have significantly advanced transparent object detection. For example, TransLab~\cite{Xie2020Segmenting} first tackled in-the-wild segmentation by combining semantic features with edge-aware losses, while polarization-enhanced methods such as Kalra et al.~\cite{Kalra2020Deep} resolve ambiguous glass boundaries through deep fusion. Light-field-based TransCut~\cite{Xu2015Transcut} segments refractive regions under complex lighting, and TransSeg~\cite{Xie2021Segmenting} applies transformer-based attention to refine thin-edge masks. Mousavi, M. et al.~\cite{Mousavi2021Supercaustics} further explore synthetic training pipelines using Supercaustics for improved real-world generalization. Beyond pure RGB segmentation, multimodal fusion and depth completion methods recover clearer geometry for transparent objects. ClearGrasp~\cite{Sajjan2020ClearGrasp} learns to predict missing depth from RGB input, enabling improved surface reconstruction that supports pose estimation. Zhu et al.~\cite{Zhu2021RGBD} propose an RGB-D implicit function to enhance surface continuity, while MvTrans~\cite{Wang2023Mvtrans} aggregates multi-view RGB for consistent segmentation across perspectives. Related works such as KeyPose~\cite{Liu2021KeyPose}, Dex-NeRF~\cite{Ichnowski2021DexNeRF}, and Evo-NeRF~\cite{Kerr2022EvoNeRF} leverage learned keypoints and radiance fields to estimate fine-grained 3D poses and structural consistency for transparent objects. Similarly, Glass-Loc~\cite{Zhou2019Glassloc} and StereoPose~\cite{Chen2023Stereopose} demonstrate how specialized light-field or stereo sensing can refine pose estimation for clear materials in cluttered or occluded scenes.

While these methods significantly advance transparent object detection and pose recovery for intact glassware and optical parts, they typically assume smooth, continuous surfaces. By contrast, transparent fragments exhibit irregular shapes, broken edges, and highly variable surface geometry, producing noisy or incomplete signals for both RGB and depth-based pipelines. Accurately recovering these fragment contours remains challenging yet essential for tasks such as sorting, damage assessment, and autonomous reassembly---motivating the need for a dedicated contour estimation framework that fuses visual and tactile cues to resolve fine-scale geometry.

\subsection{Visual–Tactile Fusion for Transparent Object Detection and Manipulation}
Recent work has demonstrated that combining vision and tactile sensing can significantly improve the detection and fine-scale manipulation of transparent objects under challenging conditions. Jiang \textit{et al.}~\cite{Jiang2021VisuoTactile} introduced a vision-guided GelSight probing strategy, where a segmentation network first identifies candidate contact regions and a high-resolution tactile sensor then captures local edge profiles to enhance surface detail recovery. Li \textit{et al.}~\cite{Li2022SimTrans12K} proposed TGCNN to jointly predict sampling points and designed a tactile calibration method and a visual–tactile fusion classifier using a TaTa gripper. Bian \textit{et al.}~\cite{Bian2023TransTouch} presented TransTouch, which uses a utility-driven policy to select sparse tactile samples; these labels are integrated via a confidence‐regularized network to reconstruct more complete surface profiles of transparent objects. Murali \textit{et al.} proposed ACTOR (Active Tactile-based Category-Level Transparent Object Reconstruction) to combine active tactile exploration with self-supervised learning for recovering 3D geometry and pose estimation for transparent objects of unknown categories~\cite{Murali2023Touch}.

These vision–tactile frameworks have demonstrated improved surface continuity and local contour estimation, enabling more reliable fine-scale manipulation of whole transparent objects. However, they remain optimized for smooth, intact geometries and fixed probing strategies, lacking the adaptive edge reconstruction and shape continuity recovery required for precise reassembly of fragmented, irregular transparent pieces.

\begin{figure*}[!t]
\centering
\includegraphics[width=0.95\textwidth]{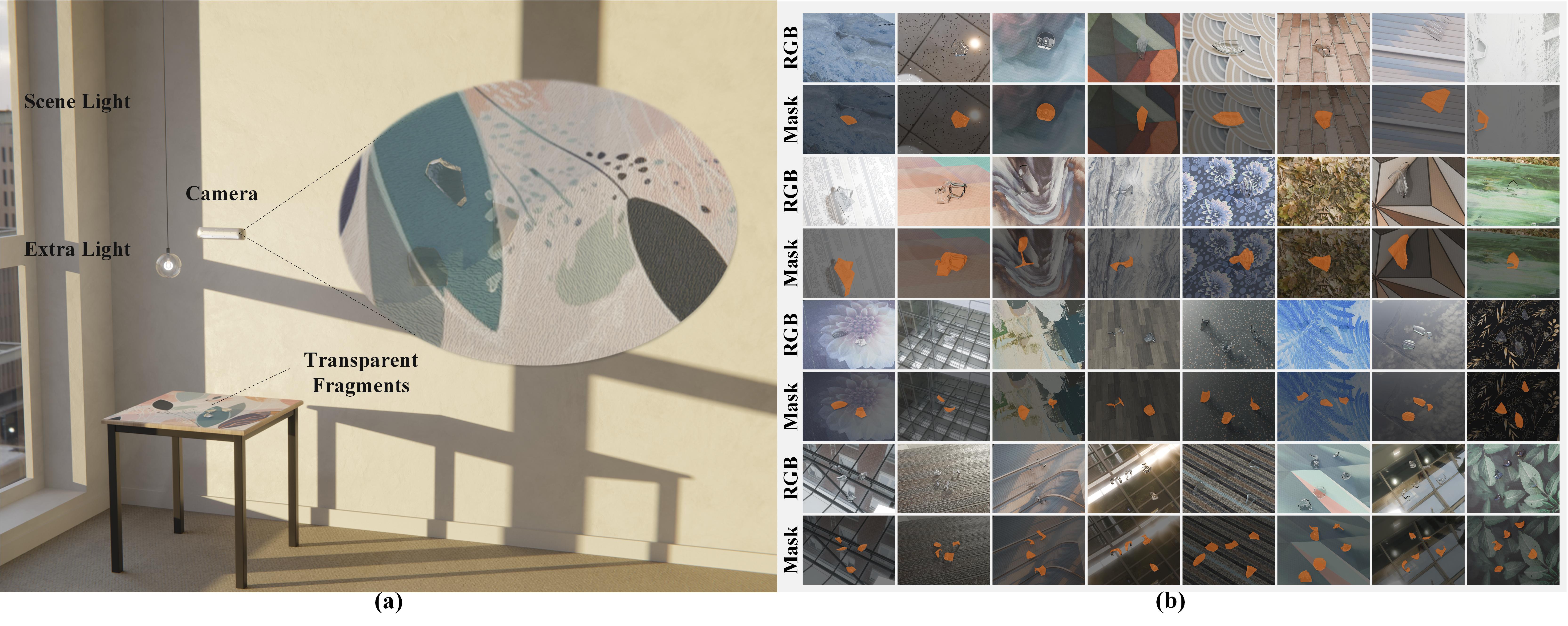}
\caption{(a) Illustrates the synthetic data generation pipeline, showcasing scene construction with elements like scene light, extra light, camera, and the projection of transparent fragments onto a surface. (b) Displays a subset of the generated dataset, featuring paired RGB images and their corresponding masks for transparent fragments.}
\label{data_augmentation}
\end{figure*}

\subsection{Fragment Reassembly}
Early unsupervised approaches relied on handcrafted geometric descriptors, such as curvature or edge features, to match fragment boundaries, but these are highly sensitive to shape ambiguities (e.g., symmetry or smooth regions) and noise, often failing when overlap is limited or edges are distorted~\cite{papaioannou2003,brown2008}.
Recent learning-based methods have significantly advanced fragment reassembly by formulating it as a graph matching or geometric reasoning task. Sellán et al.~\cite{sellan2022} introduced a benchmark dataset with a graph network to predict pairwise compatibility. DiffAssemble~\cite{scarpellini2024} unified 2D and 3D matching using graph diffusion, while Wu et al.~\cite{wu2023} used SE(3)-equivariant networks for robust transformation prediction. Other works explore joint fragment set prediction~\cite{NEURIPS2023_30ae2af8}, ~\cite{ijcai2022p146}. 

Although these approaches perform well on textured, cleanly fractured objects, they degrade on transparent fragments due to refractive distortion, irregular edges, and low-contrast masks. Our work addresses these challenges by introducing a unified visual–tactile contour estimation framework that forms the basis for robust database indexing and pairwise reassembly, enabling precise manipulation and restoration of transparent fragments.

\section{Proposed Dataset}


\begin{figure}[!t]
\centering
\includegraphics[width=3.5in]{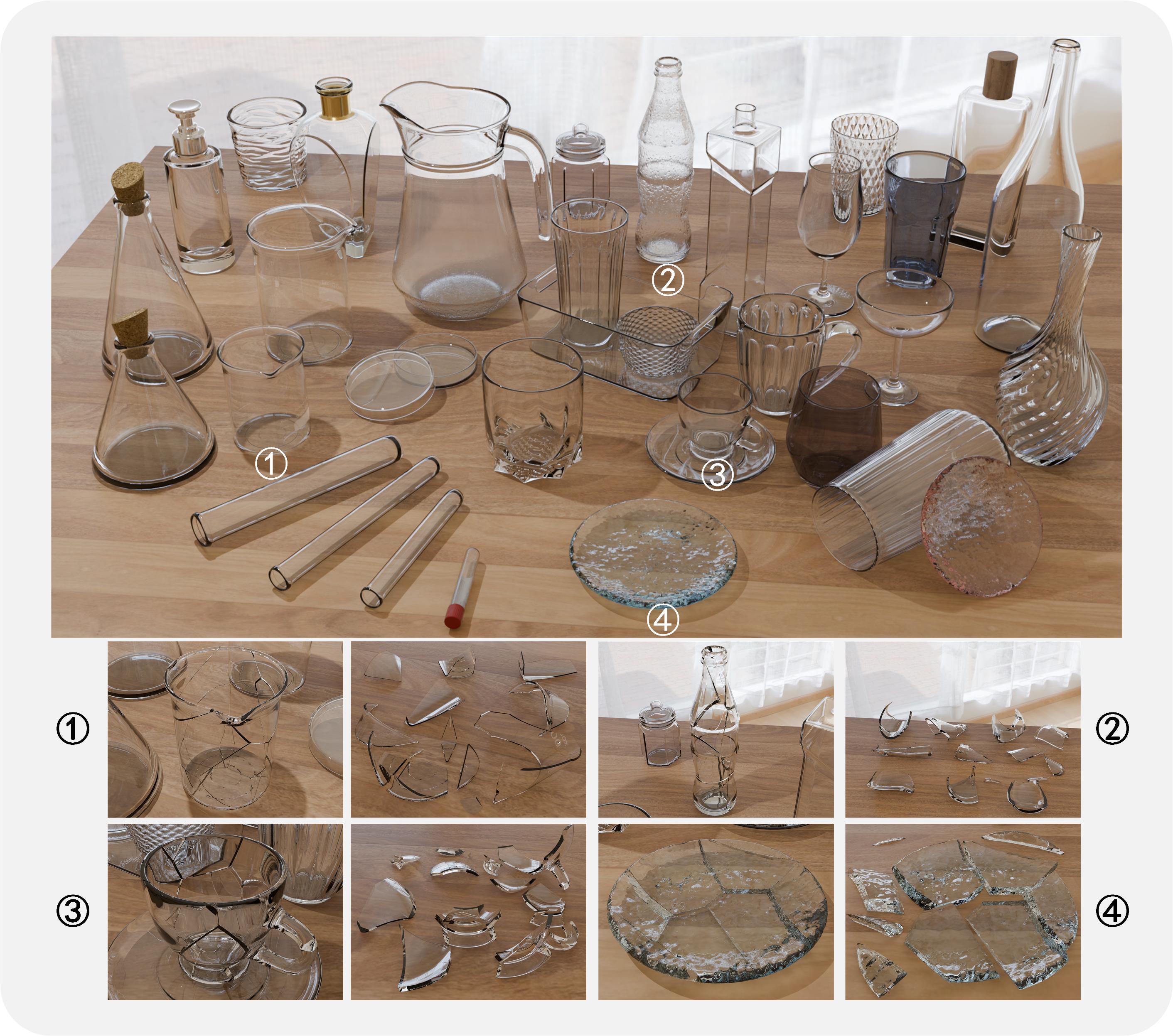}
\caption{Examples of intact transparent objects used to generate synthetic fracture scenarios in the proposed TransFrag27K dataset. The collection includes a diverse range of materials, shapes, and surface textures, such as glassware, lab beakers, cups, bottles, and decorative items. The serial number corresponds to the state of the corresponding object after adding the fracture effect.}
\label{dataset}
\end{figure}

To support research on transparent fragment perception and reassembly, we construct the first Transparent Fragment Dataset TransFrag27K which contains 27,000 synthetic and 317 real-world samples, as shown in Fig. \ref{data_augmentation} (b), and a scalable synthetic data generation pipeline. This section introduces definition and the classification method for transparent object fragments and details the dataset production process, including scene construction, rendering, and automatic annotation.

\subsection{Definition and Classification of Transparent Fragments}

In this study, we define transparent fragments as physically separable pieces resulting from mechanical breakage of desktop-scale transparent objects, such as laboratory containers, precision transparent instruments or household glassware. 

These fragments are generated by external forces and retain sufficient size and structural integrity to be manipulated by robotic grippers (typically larger than 2\,cm in their smallest dimension). Crucially, each fragment exhibits optical properties within the perceptual range of standard RGB and depth sensors. This includes materials such as clear and frosted glass, transparent plastic, crystal, and acrylic, characterized by features such as visible light transmission, refraction, specular reflection, or surface diffusion. Extremely fine shards (thickness less than 0.5\,mm), opaque or painted pieces, and fragments lacking discernible contours under normal lighting conditions are excluded from the dataset. We define a hierarchical classification method for transparent fragments along four dimensions: surface type, shape class, material class, and parent‐object context. 
Surface‐type labels distinguish fractured facets, those exhibiting irregular geometry, high curvature variance, and edge roughness typical of breakage interfaces, from intact facets, which correspond to smooth original surfaces (e.g., exterior walls or bases). 
Only fractured facets are used for matching during complete reassembly, while intact facets provide contextual cues for partial reassembly and help reject false alignments.
Next, each fragment is assigned one of three shape classes—Planar, Curved, or Irregular. A fragment is labeled Planar if at least 70\% of its surface area fits a plane within a 1mm tolerance; such shards typically originate from lids, flat bottoms, or side walls. If a fragment fails the planar check but its intact facets exhibit a consistent cylindrical or spherical curvature within a predefined mean squared fitting error, it is labeled Curved, as is common for shards from cups, bowls, and bottles. Fragments that satisfy neither condition, often due to multiple curvature modes or disconnected surfaces, such as handles or decorative rims, are labeled Irregular. This definition ensures consistency in data acquisition, supports downstream tasks such as segmentation and classification, and reflects realistic scenarios in which we interact with broken transparent objects in manufacture or lab environments, as shown in Fig. \ref{dataset}.

For practical purposes, we broadly categorize these fragments based on their material type and typical optical behavior. Glass fragments generally have high rigidity and smooth, specular surfaces; acrylic and transparent plastics tend to show slightly higher compliance and may exhibit subtle color tints under standard lighting; crystals can vary from colorless to colored and may produce more complex internal refraction. The parent object is specified by humans and is the complete object before the transparent fragments are broken. This classification provides a foundation for dataset construction, experimental setup, and the contour matching and reassembly pipeline. It ensures that the benchmark covers a diverse range of real-world fragment conditions, which is essential for validating visual–tactile fusion approaches to contour estimation and precise reassembly.
We then proposed a scalable synthetic data generation pipeline based on the definition of transparent fragments and constructed a synthetic dataset.

\subsection{Dataset Construction and Annotation}

\begin{figure*}[!t]
\centering
\includegraphics[width=0.95\textwidth]{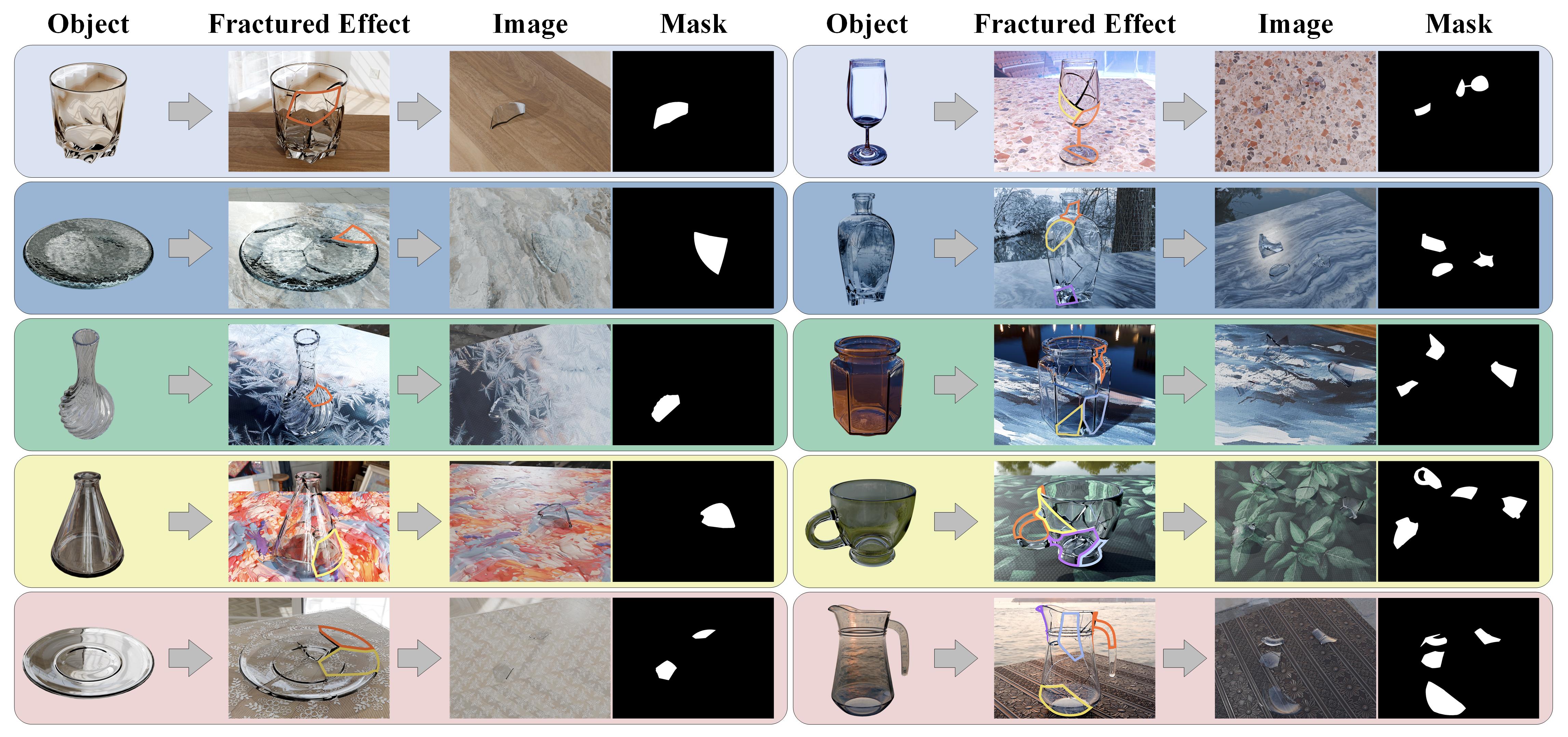}
\caption{This figure details the steps involved in generating the synthetic data for TransFrag27K. For various transparent objects (e.g., glass, plate, bottle), the second column shows the simulation of breakage, followed by the resulting RGB image and its corresponding mask, highlighting the visual shape of individual transparent fragments.}
\label{datasetshow}
\end{figure*}


Transparent fragments possess both transmission and refraction properties, making their visual appearance easily affected by lighting and background conditions. Moreover, the shapes of fractured pieces are highly irregular and unpredictable, making manual annotation in traditional dataset acquisition challenging and inefficient. To address this, we developed an automated transparent fragment dataset generation pipeline in Blender 4.2.3 \cite{Tang2023Benchmark}, using the cycles rendering engine.

First, we constructed a variety of transparent materials by adjusting parameters in a multi-layer shader node network: clear glass was implemented using a principled bidirectional scattering distribution function (BSDF) node, with an index of refraction (IOR) set to 1.5 and minimal volumetric absorption. Frosted glass and specialty surface textures were produced by raising Roughness and driving the surface normals with a procedural noise texture. Crystal materials were created using RGB-separated Refraction BSDF nodes with different IOR values to mimic dispersion, combined with high-density volumetric absorption to achieve strong refraction and edge color gradients. 
These materials were then assigned to high-resolution models, which were subsequently randomly fractured into irregular fragments using Blender’s Cell Fracture add-on.
After fracturing, the fragments are placed into the constructed scene shown in Fig. \ref{datasetshow}, which also contains a horizontal supporting plane, HDRI (i.e., high dynamic range imaging) environmental lighting, independent dynamic light sources and a camera, as shown in Fig. \ref{data_augmentation}(a). Using a parameterized script, each scene element is randomized prior to each rendering pass. The camera pose is randomized within a predefined space to capture diverse viewing angles. The planar background is replaced with textures of common surfaces, such as desktops and floors featuring diverse patterns, and is subjected to random affine transformations and displacements before rendering to enhance background diversity. Additional dynamic light sources are introduced, with type (point/area), size, intensity, and spatial position randomized within set ranges. Alongside dynamically adjusted HDRI intensity, these independent light sources combine with HDRI lighting to form varied illumination conditions, mimicking real-world lighting variability. 

Through the combination of randomized factors, the synthetic data reflects diverse and intricate real scenes. The modular, parameterized script architecture allows users to quickly generate dataset variants for different application scenarios by simply specifying parameter ranges for fundamental elements (e.g., target object occurrence frequency, average scene brightness). With manually curated asset libraries of models, materials, and HDRI textures, scenes can be rapidly assembled withoutlight model reconstruction. 
In addition to RGB images, each render also outputs a binary mask derived from the Object ID pass in Blender’s compositor nodes. The fragment masks extracted from the Object ID layer remain accurate even for regions with high transparency and faint edges, overcoming the limitations of manual annotation for transparent objects.
The resulting TransFrag27K dataset contains 27,000 images and masks at 640×480 resolution, covering over 150 background textures and 100+ HDRI maps, ensuring substantial diversity.

\section{Methodology}
\begin{figure*}[!t]
\centering
\includegraphics[width=0.95\textwidth]{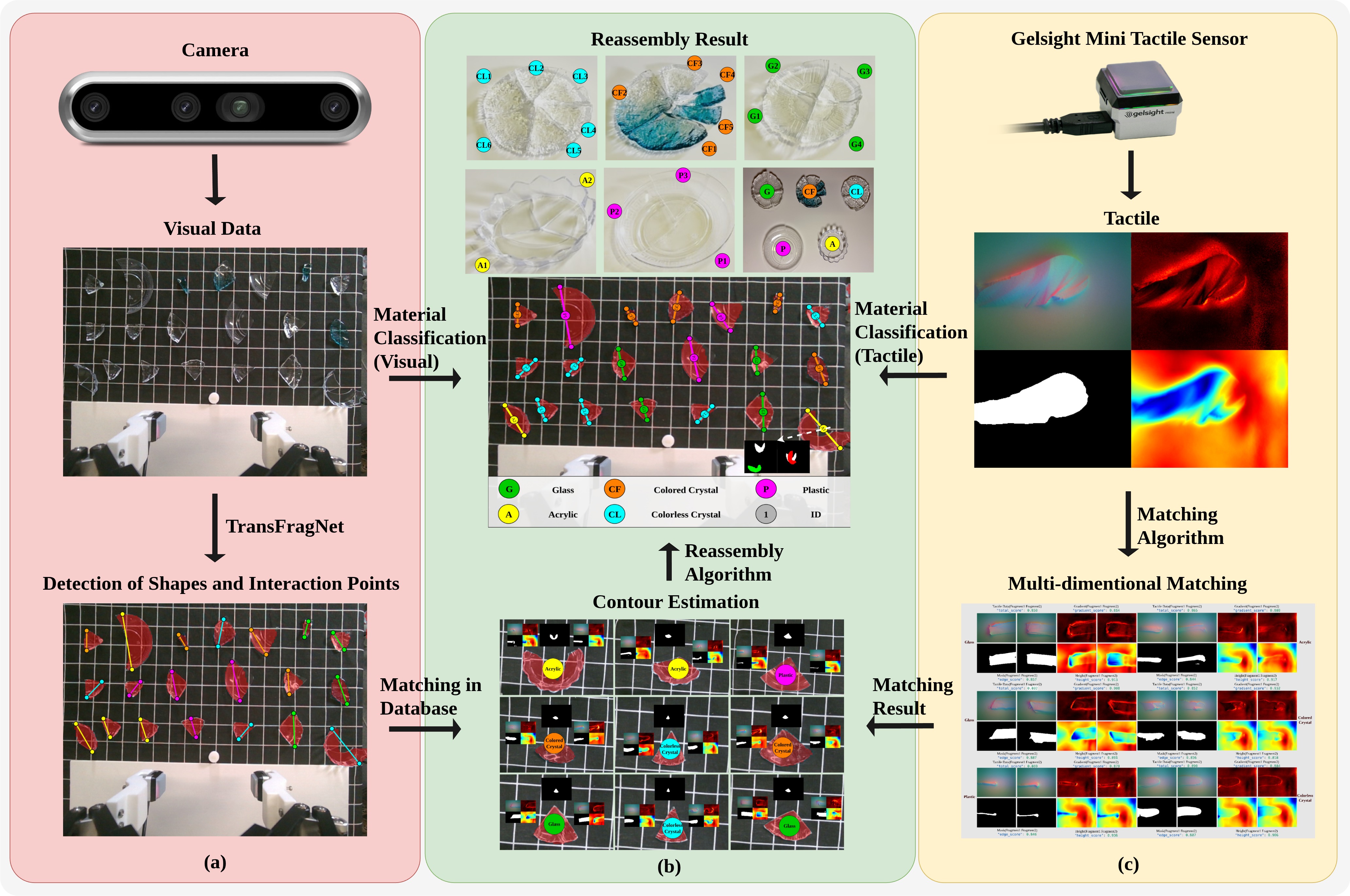}
\caption{Visual-tactile fusion based contour estimation framework for transparent fragment reassembly. (a) TransFragNet based visual grasping position detection and segmentation  (b) Fragment reassembly using visual-tactile fusion based fragments contour estimation framework in mixed fragment scenes (c) Tactile information reconstruction and matching.}
\label{finalall}
\end{figure*}

This section presents our visual–tactile fusion framework for transparent fragment contour estimation and reassembly, as illustrated in Fig.~\ref{finalall}. We first detail the core components of our contour estimation pipeline: (A) the TransFragNet for visual mask and interaction point detection, (B) the tactile reconstruction algorithm for recovering fine edge geometry, and (C) the visual–tactile fusion network for material classification. We then describe (D) how these components are integrated to produce a complete contour representation, which serves as the basis for (E) our multi-feature contour matching and reassembly algorithm.

\subsection{Transparent Fragment Grasping \& Sampling Position Detection}
\label{b-transparent-fragment-grasping-and-sampling-position-detection}


\begin{figure}[!t]
\centering
\includegraphics[width=3.5in]{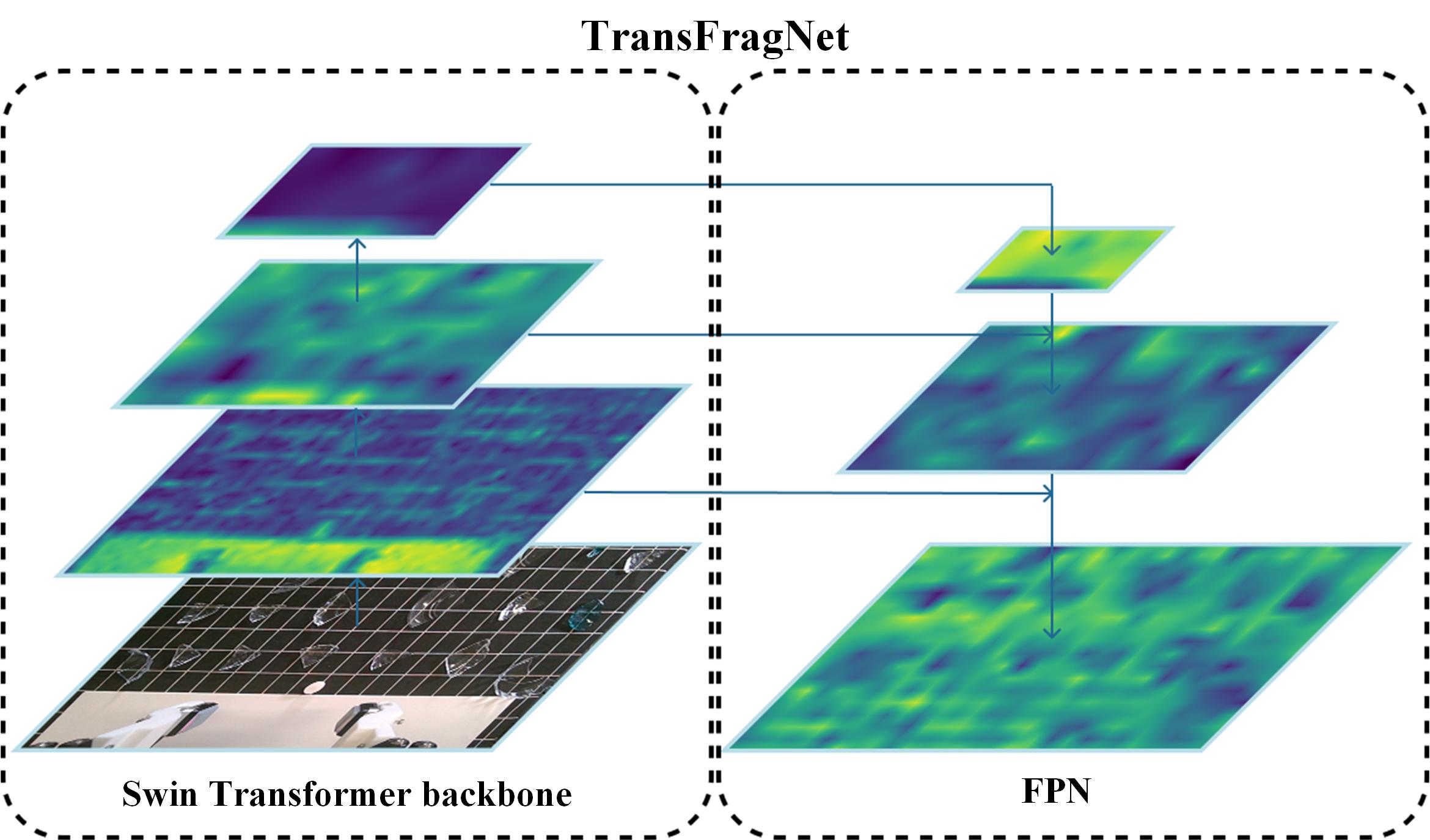}
\caption{The architecture of the proposed TransFragNet. The model employs a Swin Transformer backbone for robust feature extraction, followed by a dual-path decoder that integrates deformable transposed convolution and anti-aliased upsampling. The final segmentation head generates pixel-wise probability masks, which are further refined through morphological post-processing.}
\label{fig:STArc}
\end{figure}

\begin{figure}[!t]
\centering
\includegraphics[width=3.5in]{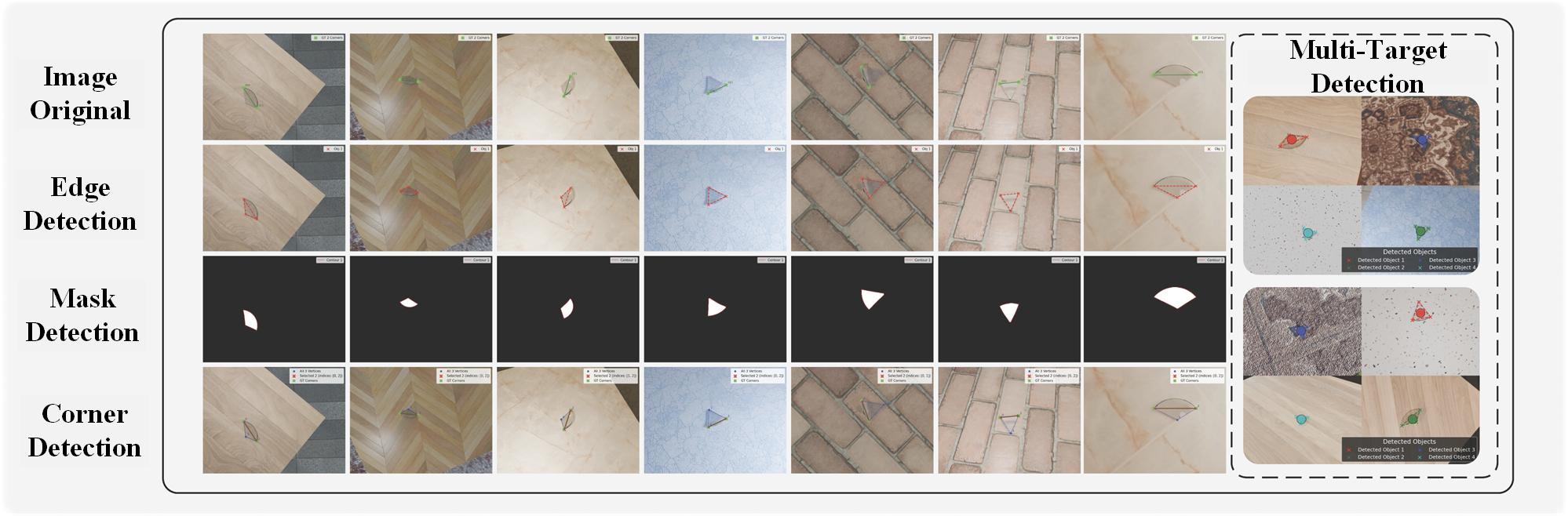}
\caption{Segmentation results for transparent fragments, including both single and multiple fragment scenarios. The top row shows the original input images, while the middle two rows display the annoted mask and predicted segmentation masks generated by our model. The bottom row illustrates the interaction corner points by applying morphological post-processing.}
\label{detection}
\end{figure}


Due to the low-contrast and ambiguous boundaries of transparent fragments and the unique optical properties, the appearance of transparent objects is easily disturbed by the backgrounds. A key challenge lies in accurately identifying these fragments from visual input and subsequently determining appropriate interaction points. 

Our approach utilizes a robust segmentation model and a precise geometric post-processing pipeline. The framework's input is an RGB image, $I$. Images are resized and normalized with standard statistics, and we employ common geometric and photometric augmentations to improve generalization under background/illumination variations and mild occlusions.
At the core of our fragment detection is a transformer encoder–decoder segmentation network (TransFragNet; see Fig.~\ref{fig:STArc}). A Swin Transformer backbone extracts multi-scale features with strong global context, capturing both low-level texture and high-level semantics that are crucial for low-contrast fragments.
Our decoder integrates components tailored for transparent fragments: an FPN-style multi-scale fusion strengthens low-contrast cues; an anti-aliased upsampling branch suppresses jagged edges on thin boundaries; and a deformable transposed-convolution branch adapts sampling to irregular, jagged contours. An attention-based edge-sharpening module further enhances boundary localization under reflection/refraction artifacts.
To balance geometric adaptability and edge fidelity, we fuse the deformable upsampling path with the anti-aliased path through a learned combination, followed by normalization/activation and edge refinement. This dual-path design preserves sharp, thin edges while remaining robust to complex, irregular geometries.
The final head uses progressive channel reduction to produce a logit map, which is converted to a probability mask $M_{pred}$ with a sigmoid:
\begin{equation}
M_{pred}(x,y) = \sigma\!\big( (f_{dec} \circ f_{enc})(I; \theta_{model})(x,y) \big).
\end{equation}
Lightweight post-processing (smoothing + morphology) removes small holes/isolated noise and stabilizes contours without altering fine boundaries. The model is trained using the Tversky loss function, which provides better performance in imbalanced segmentation tasks compared to the standard binary cross-entropy loss. The Tversky loss is defined as:
\begin{equation}
L_{Tversky} = 1 - \frac{TP}{TP + \alpha FN + \beta FP},
\end{equation}
where $TP$, $FN$, and $FP$ represent True Positives, False Negatives, and False Positives, respectively. We use $\alpha = 0.7$ and $\beta = 0.3$ to emphasize recall over precision in the context of transparent fragment detection.

During inference, we adopt a minimal Test-Time Augmentation (TTA) \cite{Shanmugam2021} (flip/scale/rotate) and average predictions for robustness across backgrounds/lighting, avoiding parameter-heavy ensembles.
For extracting interaction points, we apply a geometric post-processing algorithm on the predicted mask: contours are found per fragment, and the endpoints of the longest chord (principal axis proxy) are selected as grasping/sampling locations. This avoids concavity traps, aligns with stable support directions, and extends naturally to multiple-object scenes via simple area filtering. These interaction points are then sent to the robot to guide targeted tactile sampling.
We evaluate with mask IoU (i.e., Intersection-over-Union) and corner distance error. On our test set covering diverse materials and lighting, the detector attains a mean IoU of 0.8818 and yields stable interaction points for reliable tactile sampling (as shown in Fig.~\ref{detection}). Compared to traditional segmentation methods, our approach demonstrates superior robustness in handling the low-contrast and ambiguous boundaries of transparent fragments. The integration of advanced data augmentation techniques and the Swin Transformer backbone ensures that our model generalizes effectively across diverse scenarios, as validated by its significantly higher IoU scores in both synthetic and real-world data.

\subsection{Tactile Reconstruction Algorithm}
\label{Subsection: IV.C}

\begin{figure}[!t]
\centering
\includegraphics[width=3.5in]{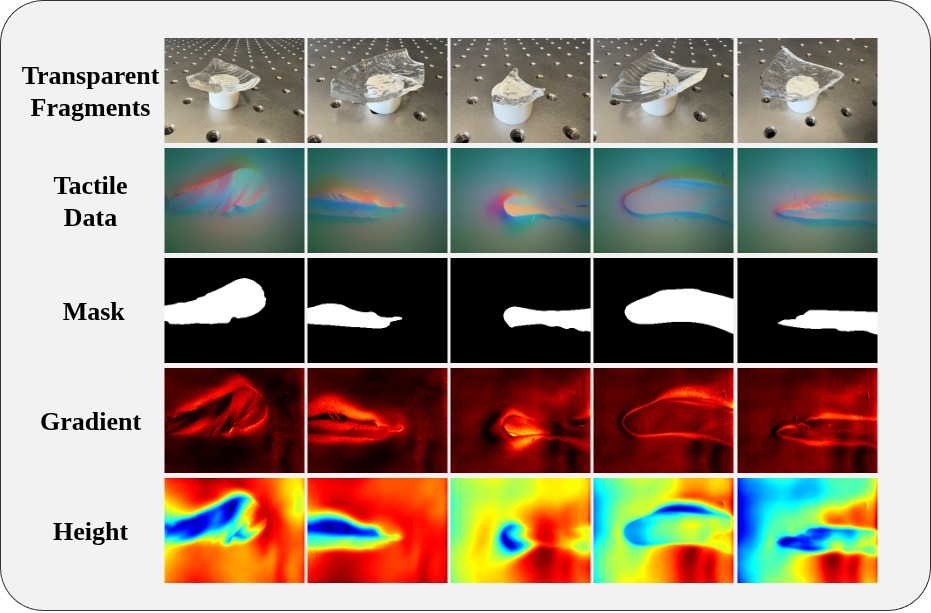}
\caption{Tactile reconstruction pipeline: recovering contact masks, local gradient fields, and detailed height maps from GelSight Mini data. These outputs provide fine-scale edge and curvature details critical for contour estimation and fragment matching.}
\label{fig:reconstruct}
\end{figure}


The tactile reconstruction pipeline converts raw GelSight Mini images into precise local geometry representations, including contact masks, gradient fields, and height maps, as illustrated in Fig.~\ref{fig:reconstruct}.  
By recovering both the 2D contact region and the underlying 3D surface profile, we obtain the fine‐scale edge and curvature cues that are invisible to conventional depth cameras.

We first segment the contact region in each GelSight Mini image using a U-Net based convolutional network, as shown in Fig.~\ref{fig:unet}. The encoder consists of four blocks, each with two $3\times3$ convolutions followed by max‐pooling, with channel depths of 64, 128, 256, and 512. The decoder mirrors this structure via transposed convolutions and skip‐connections that restore spatial detail. We train the network with a combined binary cross‐entropy and Dice loss to sharpen mask boundaries. In our experiments, this model achieves an average IoU of 0.967 on real-world tactile frames, reliably isolating the fragment’s contact footprint for subsequent processing. 
Once the contact region mask \(M(x,y)\) is available, we recover the local surface geometry in two steps: gradient estimation via photometric stereo, and height reconstruction via a frequency-domain Poisson solver formulated using the Discrete Cosine Transform (DCT) \cite{Wang2021Gelsight}:

\begin{figure}[!t]
\centering
\includegraphics[width=3.5in]{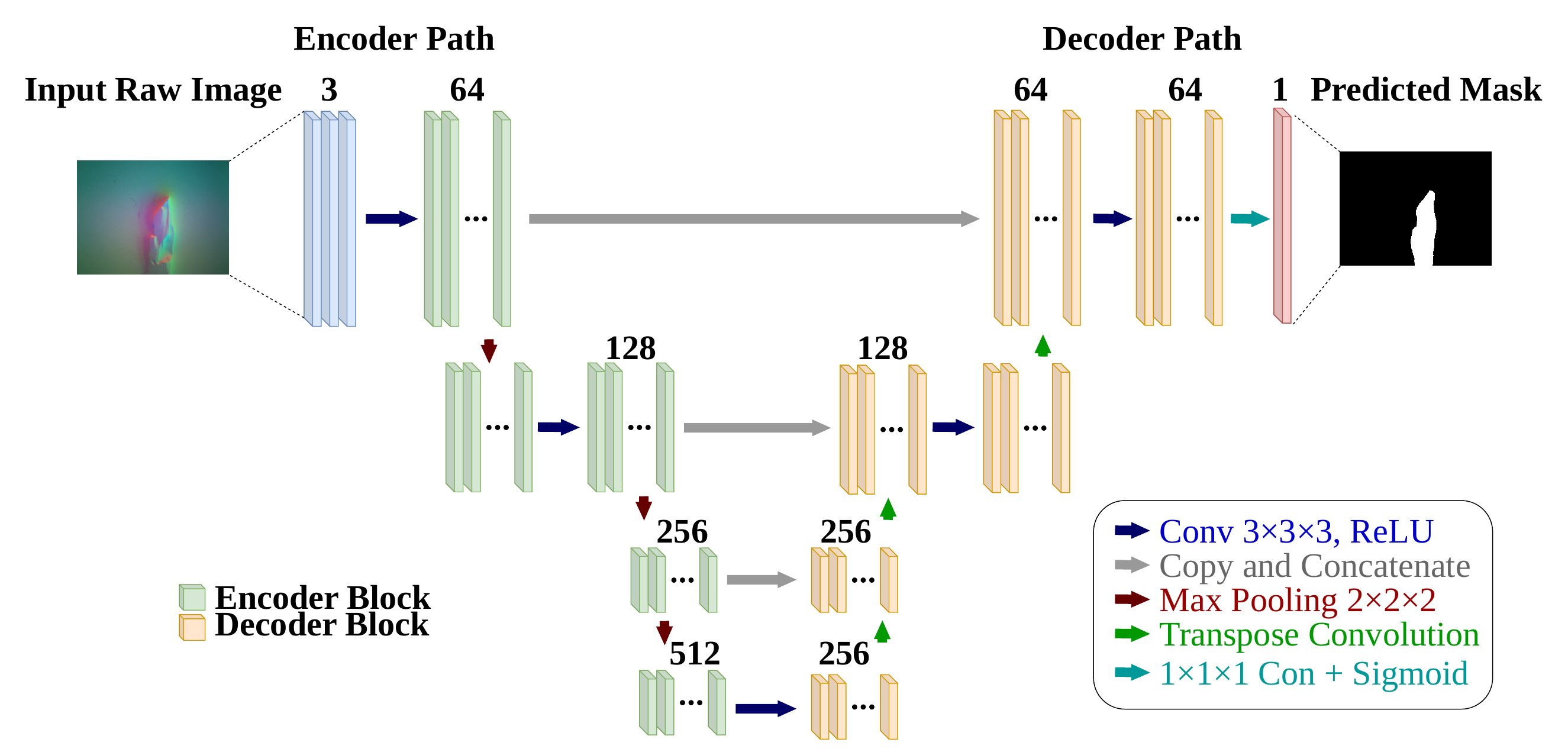}
\caption{Architecture of the U-Net convolutional network used for segmenting tactile contact regions. This architecture is specifically designed to segment the contact regions in GelSight Mini tactile images, achieving high accuracy and sharp boundary delineation for subsequent tactile reconstruction tasks.}
\label{fig:unet}
\end{figure}

First, Given the segmented GelSight Mini image channels \(I_r,I_g,I_b\) and known light directions \(\{\mathbf{s}_k\}_{k=1}^3\), we solve at each contact pixel \((x,y)\) the linear system
\begin{equation}
\begin{bmatrix}
\mathbf{s}_1^\top\\
\mathbf{s}_2^\top\\
\mathbf{s}_3^\top
\end{bmatrix}
\mathbf{n}(x,y)
=
\begin{bmatrix}
I_r(x,y)\\
I_g(x,y)\\
I_b(x,y)
\end{bmatrix},
\end{equation}
in the least-squares sense for the surface normal \(\mathbf{n}=(n_x,n_y,n_z)^\top\).  The two‐dimensional gradient field \(\mathbf{G}=(G_x,G_y)\) is then
\begin{equation}
G_x = -\frac{n_x}{n_z}, 
\quad
G_y = -\frac{n_y}{n_z},
\end{equation}
defined for all \((x,y)\) where \(M(x,y)=1\). 
Second, To achieve robust, high-fidelity surface recovery, especially in regions with sharp discontinuities or measurement noise, we adopt a frequency-domain Poisson solver with Neumann boundary conditions using the Discrete Cosine Transform (DCT). Given the divergence field  
\begin{equation}
f(x,y) = \frac{\partial G_x}{\partial x} + \frac{\partial G_y}{\partial y},
\end{equation}
we modify $f$ near the domain boundaries to impose consistent Neumann boundary conditions derived from the input gradients. The resulting Poisson equation is solved in the DCT domain:
\begin{equation}
\hat{H}(u,v) = -\frac{\hat{f}(u,v)}{\lambda(u,v)},
\end{equation}
where $\hat{f}(u,v)$ is the DCT of $f$ and $\lambda(u,v)$ denotes the discrete Laplacian eigenvalues. An inverse DCT then yields the final height map $H(x,y)$. This frequency-domain formulation provides a fast, closed-form solution with subpixel accuracy, enabling reliable contour matching and reassembly.

This reconstruction algorithm produces high-fidelity contact height profiles that capture fine edge and curvature details of the transparent fragments, forming the basis for robust contour estimation and reassembly.

\subsection{Visual-Tactile Fusion Material Classification}

\begin{figure}[!t]
\centering
\includegraphics[width=3.5in]{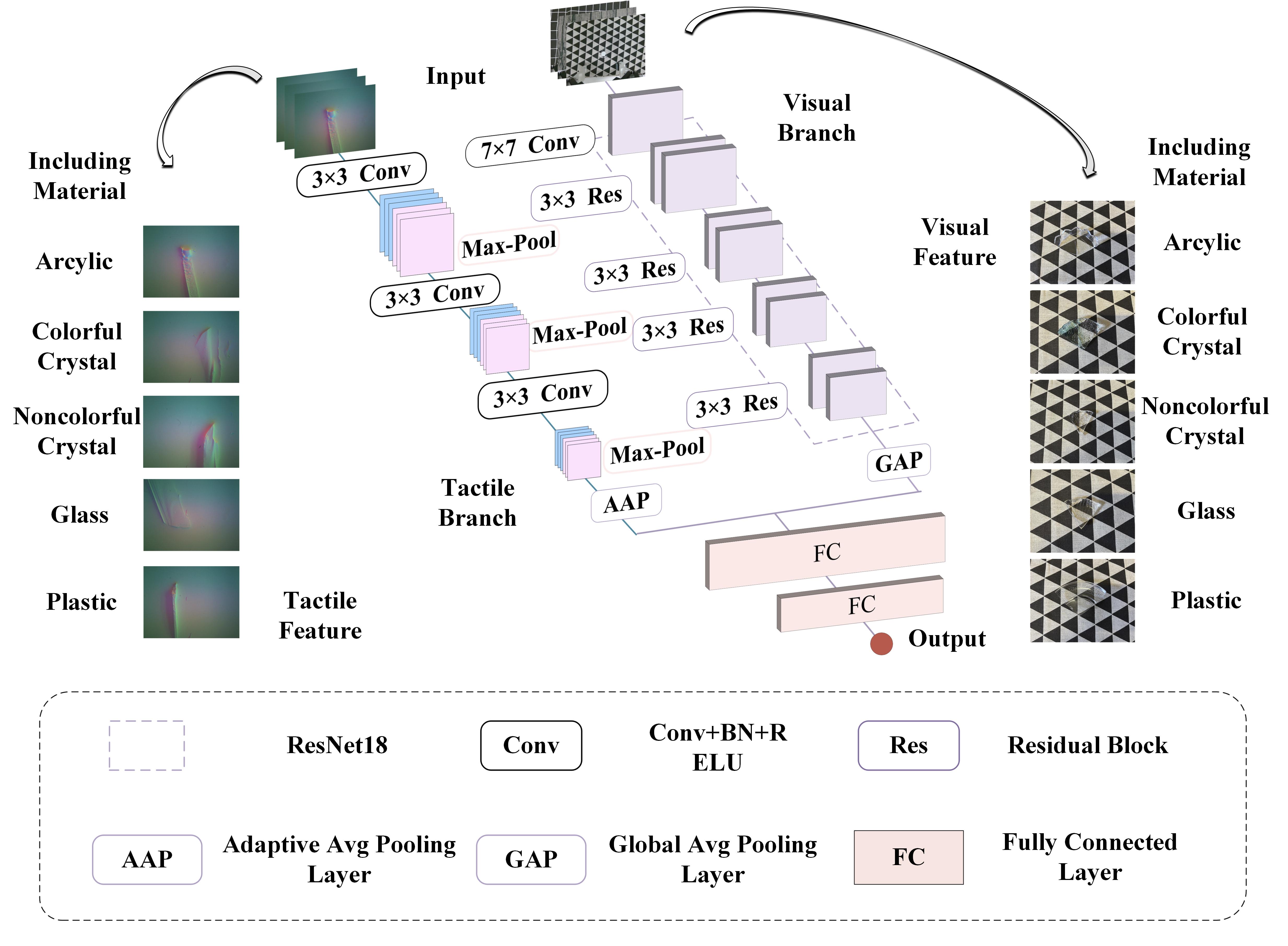}
\caption{Dual-branch late fusion network architecture for visual-tactile material classification. The visual branch employs ResNet-18 for robust feature extraction from RGB images, while the tactile branch uses a lightweight CNN to process GelSight Mini tactile data. Features are concatenated at the late fusion stage and processed through a multi-layer classifier.}
\label{fig:fusion_network}
\end{figure}

Accurate material classification of transparent fragments is crucial for effective reassembly, as different materials (glass, plastic, acrylic, colored crystal, and colorless crystal) exhibit distinct optical and tactile properties that affect both manipulation strategies and reassembly compatibility. 

While visual-only approaches often fail due to refractive distortion and low contrast boundaries in transparent objects, the integration of tactile sensing provides complementary material-specific cues that significantly enhance classification robustness. Therefore, Our approach leverages a dual-branch late fusion architecture that combines global visual context with local tactile texture information. This design philosophy acknowledges that visual data provides rich contextual information about overall shape and appearance, while tactile data captures fine-scale surface properties that are imperceptible to vision-based sensors. Given the substantial difference in data availability between modalities, 300 visual samples versus 100 tactile samples across five material categories. We adopt a visual-dominant fusion strategy where tactile information serves as an auxiliary modality to refine material discrimination.

Our dual-branch fusion network, as shown in Fig. \ref{fig:fusion_network}, consists of three primary components: a visual feature extraction branch, a tactile feature extraction branch, and a late fusion classifier.
The visual branch employs a ResNet-18 backbone pre-trained on ImageNet, providing robust feature extraction capabilities while maintaining computational efficiency. Given an input RGB image $I_v \in \mathbb{R}^{3 \times 224 \times 224}$, the visual backbone extracts hierarchical features through residual connections:
\begin{equation}
F_v = f_{visual}(I_v; \theta_v) \in \mathbb{R}^{512},
\end{equation}
where $f_{visual}$ represents the ResNet-18 encoder and $\theta_v$ denotes the trainable parameters. The final fully connected layer is replaced with a linear projection to map features to a 512-dimensional visual feature space, enabling effective transfer learning from natural image domains to transparent material classification.
For tactile feature extraction, we design a lightweight CNN architecture to process GelSight Mini tactile images $I_t \in \mathbb{R}^{3 \times 128 \times 128}$. The architecture prioritizes spatial efficiency and regularization to prevent overfitting on limited tactile data:
\begin{equation}
F_t = f_{tactile}(I_t; \theta_t) \in \mathbb{R}^{128}.
\end{equation}
The tactile branch consists of three convolutional blocks, each containing $3 \times 3$ convolutions, ReLU activations, max-pooling, and batch normalization. Channel dimensions progress as $3 \rightarrow 16 \rightarrow 32 \rightarrow 64$, with adaptive global average pooling reducing spatial dimensions to $1 \times 1$ before a final linear projection to 128-dimensional tactile features. This compact representation captures essential material texture and surface roughness cues while maintaining computational efficiency.
Rather than early or intermediate fusion approaches that might amplify noise in the limited tactile data, we employ late fusion to combine modality-specific features:
\begin{equation}
F_{fused} = [F_v; F_t] \in \mathbb{R}^{640},
\end{equation}
where $[·; ·]$ denotes concatenation. The fused features are processed through a multi-layer classifier with dropout regularization:
\begin{equation}
\hat{y} = f_{classifier}(F_{fused}; \theta_c),
\end{equation}
where the classifier consists of three fully connected layers ($640 \rightarrow 256 \rightarrow 128 \rightarrow 5$) with ReLU activations and dropout rates of 0.5 and 0.3, respectively. This architecture balances representation capacity with regularization to handle the limited training data effectively.

To address the imbalance between modalities, we implement carefully designed augmentation strategies. For visual data, we apply standard augmentations including random horizontal flips (p=0.5), rotations (±15°), color jittering (brightness±0.3, contrast±0.3, saturation±0.2, hue±0.1), and random affine transformations. For tactile data, augmentations are more conservative to preserve physical consistency: horizontal flips (p=0.5), limited rotations (±10°), and restricted color adjustments (brightness±0.2, contrast±0.2).
We employ standard cross-entropy loss for the 5-class classification task:
\begin{equation}
\mathcal{L} = -\frac{1}{N}\sum_{i=1}^{N}\sum_{c=1}^{C} y_{i,c} \log(\hat{y}_{i,c}),
\end{equation}
where $N$ is the batch size, $C=5$ is the number of classes, $y_{i,c}$ is the ground truth label, and $\hat{y}_{i,c}$ is the predicted probability. The network is optimized using Adam with a learning rate of $1 \times 10^{-4}$, weight decay of $1 \times 10^{-4}$, and a ReduceLROnPlateau scheduler that halves the learning rate when validation loss plateaus.

By leveraging the complementary strengths of visual and tactile modalities, our dual-branch late fusion network achieves significantly higher classification accuracy compared to visual-only or tactile-only approaches. This fusion strategy not only enhances robustness against visual distortions but also provides critical disambiguation for visually similar materials, as demonstrated by its superior performance across all five material categories. This material classification framework provides essential information for the subsequent reassembly process, where material compatibility and surface properties inform contour estimation, matching and reassembly strategies.

\subsection{Visual–Tactile Contour Estimation}
\label{Subsection: IV.D}

\begin{figure}[!t]
\centering
\includegraphics[width=3.4in]{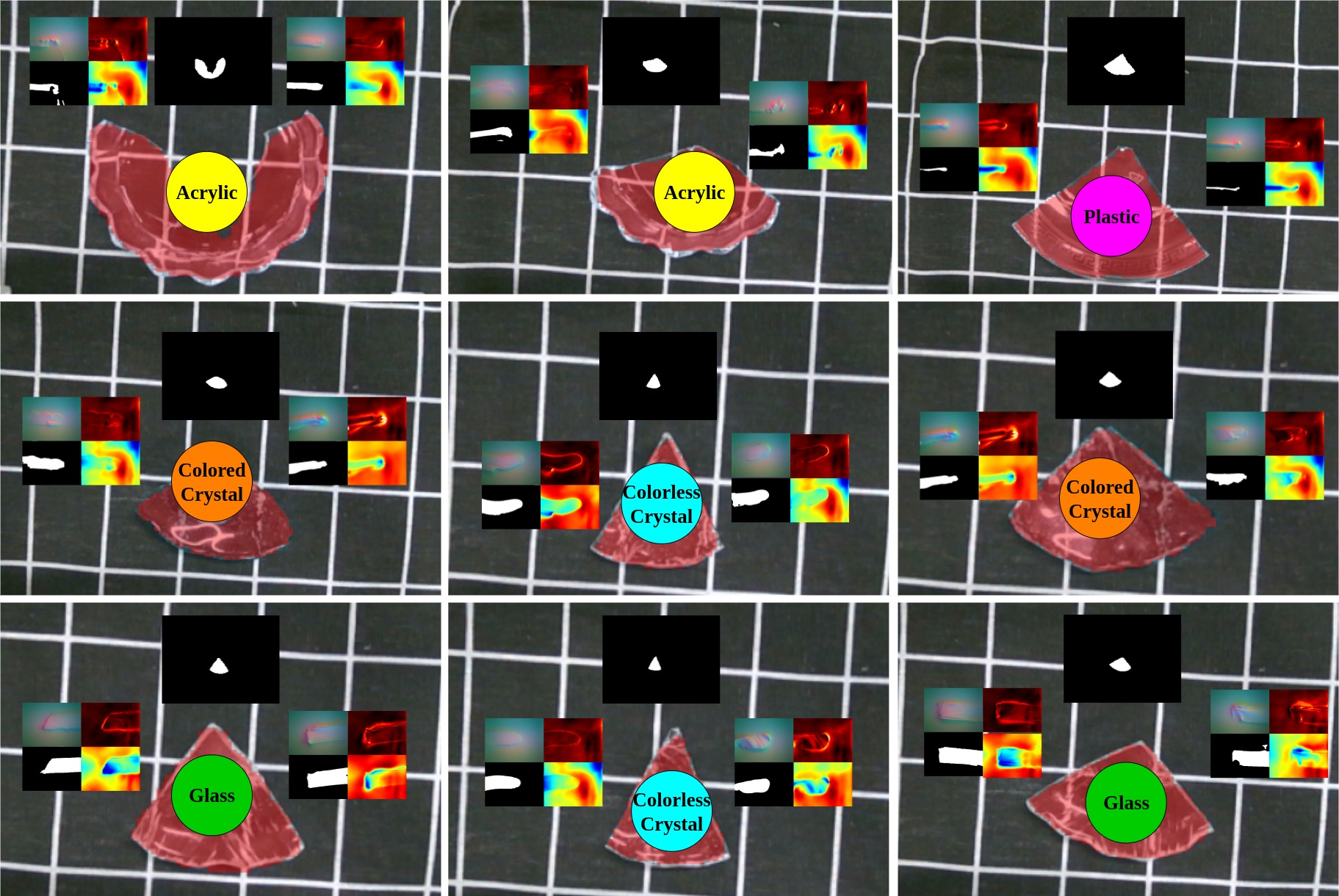}
\caption{Visual–tactile contour estimation results for transparent fragments. Each panel shows an individual fragment with its extracted global mask (center top) and reconstructed tactile profiles: RGB tactile image, edge mask, gradient map, and height map. The color-coded label indicates the final material classification result obtained through visual–tactile fusion.}
\label{fig:contour}
\end{figure}

Building on the visual detection, tactile sampling, and material classification stages described above, we combine these complementary signals into a unified contour estimation process that provides each fragment’s complete geometric representation for reassembly, as shown in Fig. \ref{fig:contour}.

First, a coarse global mask is extracted using our trained visual detection network, offering an initial outline of each fragment’s external shape. To recover fine local details that may be lost due to transparency or occlusions, we perform targeted edge sampling with Gelsight Mini tactile sensors mounted on the robotic gripper. Raw tactile images are then reconstructed into high-resolution height and edge profiles, which capture micro features that enhance the visual mask. The global visual mask and local tactile profiles are fused by aligning the tactile contour with the visual contour via centroid registration and local rigid transformation. This fusion yields a refined, complete contour that combines global geometry with local surface fidelity. 
The visual contour serves as a robust index for querying our fragment database: each estimated visual contour is matched to stored entries containing visual information, pre-sampled tactile edge, height maps, local gradient distributions, and material labels obtained during the initial visual–tactile classification.
This database-guided indexing bridges directly to the subsequent contour matching and reassembly process, where the retrieved tactile and material data enable the edge geometry, region-gradient, and height-extrema based contour matching strategies to be applied for accurate, autonomous reassembly.

\subsection{Contour Matching and Reassembly Algorithm}

\begin{figure}[!t]
\centering
\includegraphics[width=3.5in]{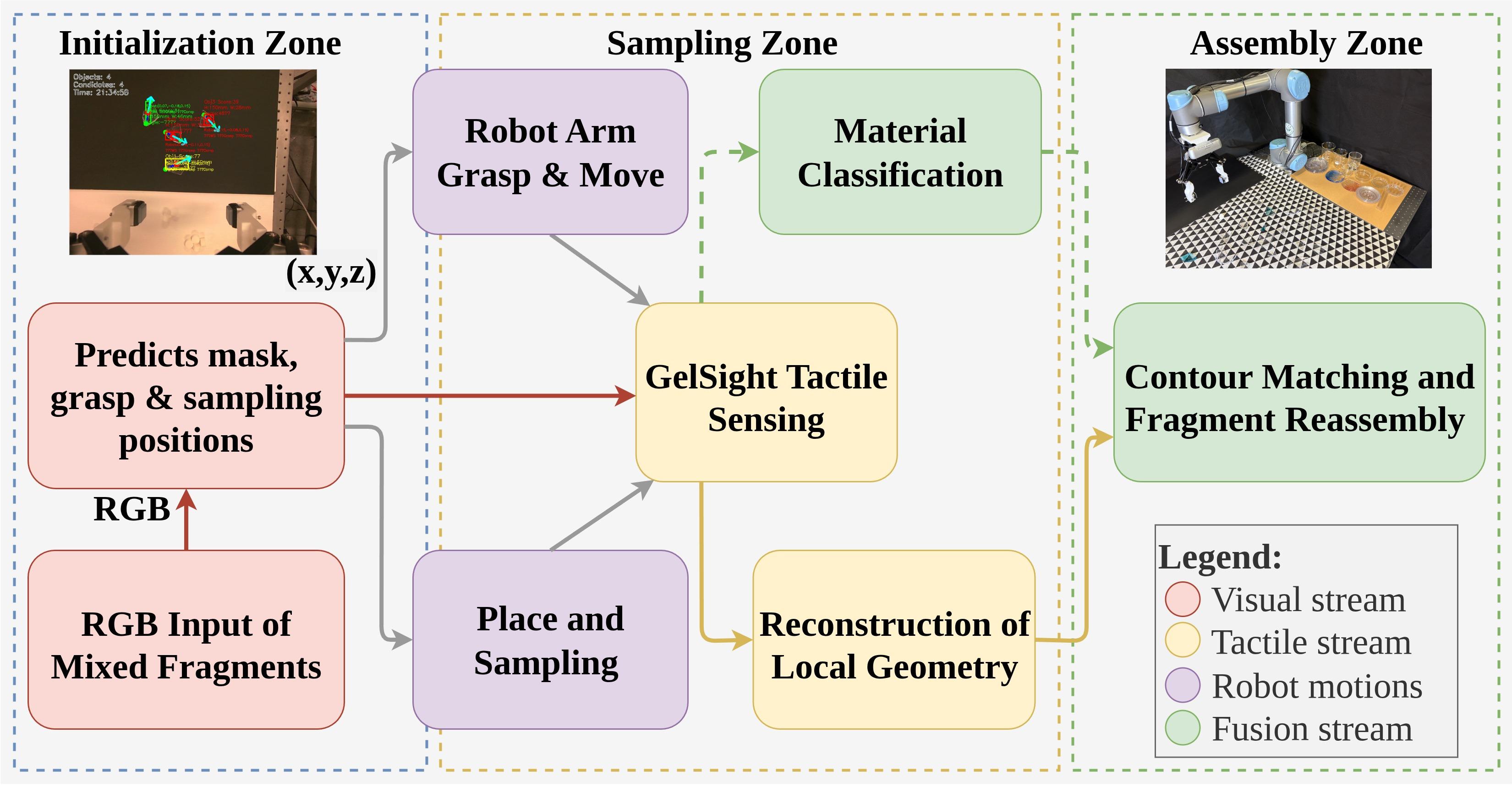}
\caption{Framework overview of the proposed visual-tactile fusion based contour estimation for transparent fragment reassembly.}
\label{fig:pipeline}
\end{figure}

To support autonomous reassembly of transparent objects under varying fracture conditions, we propose a fusion framework that combines visual detection, tactile reconstruction, and contour matching within a general contour estimation framework, as shown in Fig. \ref{fig:pipeline}.

First, each fragment’s global shape is obtained by extracting a visual mask using our trained visual detection network. For more detailed surface information, robotic GelSight Mini-based edge sampling is performed, reconstructing local edge, height and gradient maps that capture fine fracture surface features. Visual and tactile cues are then fused into a detailed contour representation used as the basis for database indexing and matching. The fragment database stores (1) RGB images and visual masks, (2) reconstructed tactile profiles, including edge gradients and height-extrema, and (3) a material label determined by visual–tactile fusion classification. At runtime, a fragment's mask is matched to database entries using IoU and Chamfer distance. 
To evaluate our visual shape-based reassembly, we report Top-1 Accuracy (the percentage of fragments for which the database index predicted by the visual matching exactly matches the ground truth), Top-3 Accuracy (whether the correct index appears within the top three candidates), and the mean Intersection-over-Union (IoU) between aligned fragment masks and their target regions. If multiple fragments have highly similar shapes or materials, we define a clear fallback condition: 
tactile re-sampling and material classification are triggered when the difference between the Top-1 score and the scores of the Top-2 or Top-3 candidates is less than ±5\% or when the maximum IoU falls below a reliability threshold (e.g., 0.70). In these cases, we will resample and reconstruct tactile height maps, edge profiles, and region gradients to refine the fragment's database index and improve matching confidence. Specifically, the robot performs real-time tactile sampling of the fragment, reconstructs its tactile features, and re-applies material classification to assign a semantic label. This restricts the candidate set to fragments of the same material class, avoiding cross-material confusion. Matching then proceeds using the standard tactile feature extraction and fusion strategy, ensuring robustness even in previously unseen object scenarios.

\begin{figure}[!t]
\centering
\includegraphics[width=3.4in]{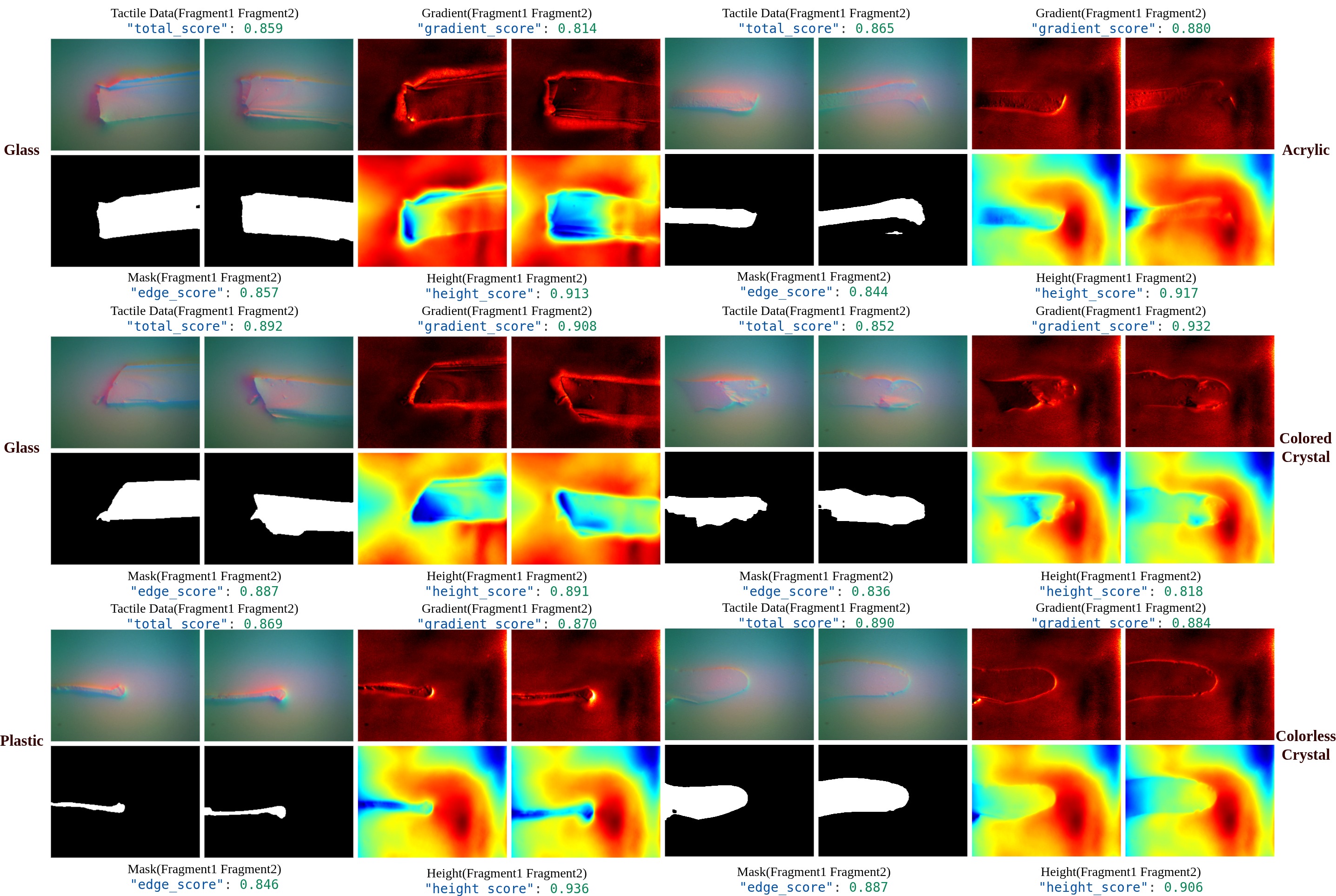}
\caption{The contour matching results and scores of fragments of some different materials: including edge geometry alignment, local area gradient matching, and height extreme value complementation, are combined into a unified score for robust transparent fragment reassembly.}
\label{fig:edgematch}
\end{figure}

The matching and reassembly process first flips the first tactile image along its midline to ensure orientation correspondence, and then matches the fragment pairs using a three-stage alignment strategy that combines geometric and tactile cues, as shown in \ref{fig:edgematch}. The alignment is evaluated using the following scoring metrics:
  
First, Given boundary point sets \(E_1,E_2\) derived from tactile contact masks and a candidate rigid transform \(T\), we compute the chamfer‐based edge score
\begin{equation}
s_{\rm edge}(T)
= \exp\Bigl(-\tfrac{1}{|E_1|}\sum_{p\in E_1}\min_{q\in T(E_2)}\|p-q\|\Bigr),
\end{equation}
which rewards tight alignment of fragment contours.
 
Next, we verify local gradient alignment by comparing orientation distributions in small windows around each edge point. For each \(p\in E_1\) and its corresponding \(T(p)\in E_2\), extract the 16‐bin orientation histograms \(\mathbf{h}_1,\mathbf{h}_2\) and dominant angles \(\phi^*_1,\phi^*_2\).  The local gradient score is
\begin{equation}
s_{\rm grad}(p, T(p)) =
\exp\left(
  -\frac{|\phi^*_1 - \phi^*_2|}{\sigma_\phi}
  -\frac{\|\mathbf{h}_1 - \mathbf{h}_2\|_2}{\sigma_h}
\right).
\end{equation}
Averaging over all boundary samples yields
\begin{equation}
S_{\rm grad}(T) = \frac{1}{|E_1|}\sum_{p\in E_1}s_{\rm grad}(p,T(p)).
\end{equation}

Finally, we exploit physical complementarity on the fracture surfaces by matching peaks to troughs in the tactile height maps. Let \(\{(y_{\max}^i,x_{\max}^i)\}\) and \(\{(y_{\min}^i,x_{\min}^i)\}\) be the local maxima/minima in height maps \(H_i\).  After applying \(T\) to the fragment under matching, we compute the complementary distance
\begin{equation}
d = \min\bigl(\|p_{\max}^1 - T(p_{\min}^2)\|,\;\|p_{\min}^1 - T(p_{\max}^2)\|\bigr),
\end{equation}
and the height‐range ratio
\(\rho = \min(r_1,r_2)/\max(r_1,r_2)\).  The extrema score is
\begin{equation}
S_{\rm height}(T)
= \frac{\alpha}{1 + d/\delta} \;+\; (1-\alpha)\,\rho.
\end{equation}

We combine these three scores into a unified metric that balances global shape, local texture, and physical interlock. The overall matching score for transform \(T\) is a weighted sum:
\begin{equation}
S(T) = w_e\,s_{\rm edge}(T) \;+\; w_g\,S_{\rm grad}(T)\;+\; w_h\,S_{\rm height}(T),
\end{equation}
with \(w_e{+}w_g{+}w_h=1\).  We rank candidate transforms by \(S(T)\) and execute the highest‐scoring alignment for fragment reassembly.

If the object is only partially fractured, where the main body is largely intact but has one or two missing pieces, we employ a vision-only shape matching method. In this special case, the missing region is segmented as a gap using the Segment Anything Model (SAM) \cite{Kirillov2023} and matched with candidate fragments based on mask alignment alone, without additional tactile sampling. 
Our contour matching and reassembly strategy, which combines edge geometry, region-gradient, and height-extrema cues, ensures robust alignment even for geometrically ambiguous fragments. This comprehensive approach addresses the limitations of single-cue methods, achieving consistently high matching accuracy across diverse materials and fracture patterns, as validated in our experiments.

\section{Experiments}

This section describes the experimental setup and results for each part of the proposed visual–tactile transparent fragment reassembly framework. First, to evaluate the constructed TransFrag27K dataset and the effectiveness of the TransFragNet detection network, we conducted detection experiments on synthetic transparent fragments (Experiment 1), real transparent fragments under varying backgrounds (Experiment 2), and robustness tests under different illumination intensities (Experiment 3).
Second, to validate the contour estimation and matching strategy under different fracture conditions, we designed two levels of reassembly tasks. We first tested partially fractured objects that maintain an approximate global shape but have one or two detached fragments (Experiment 4). In this scenario, we used a vision-only contour matching method to align detached fragments to the missing gaps. Next, we addressed completely fractured objects with no remaining global structure (Experiment 5). Here, the framework uses visual detection, tactile sampling, and contour estimation in combination with contour matching to reassemble the fragments. Finally, we evaluated the framework’s robustness and generalization in mixed scenes containing fragments from multiple materials and similar shapes (Experiment 6), verifying whether the framework can handle unknown or visually similar fragments.
We conclude this section by discussing observed limitations of the proposed framework.


\subsection{Hardware Setup}

\begin{figure}[!t]
\centering
\includegraphics[width=3.5in]{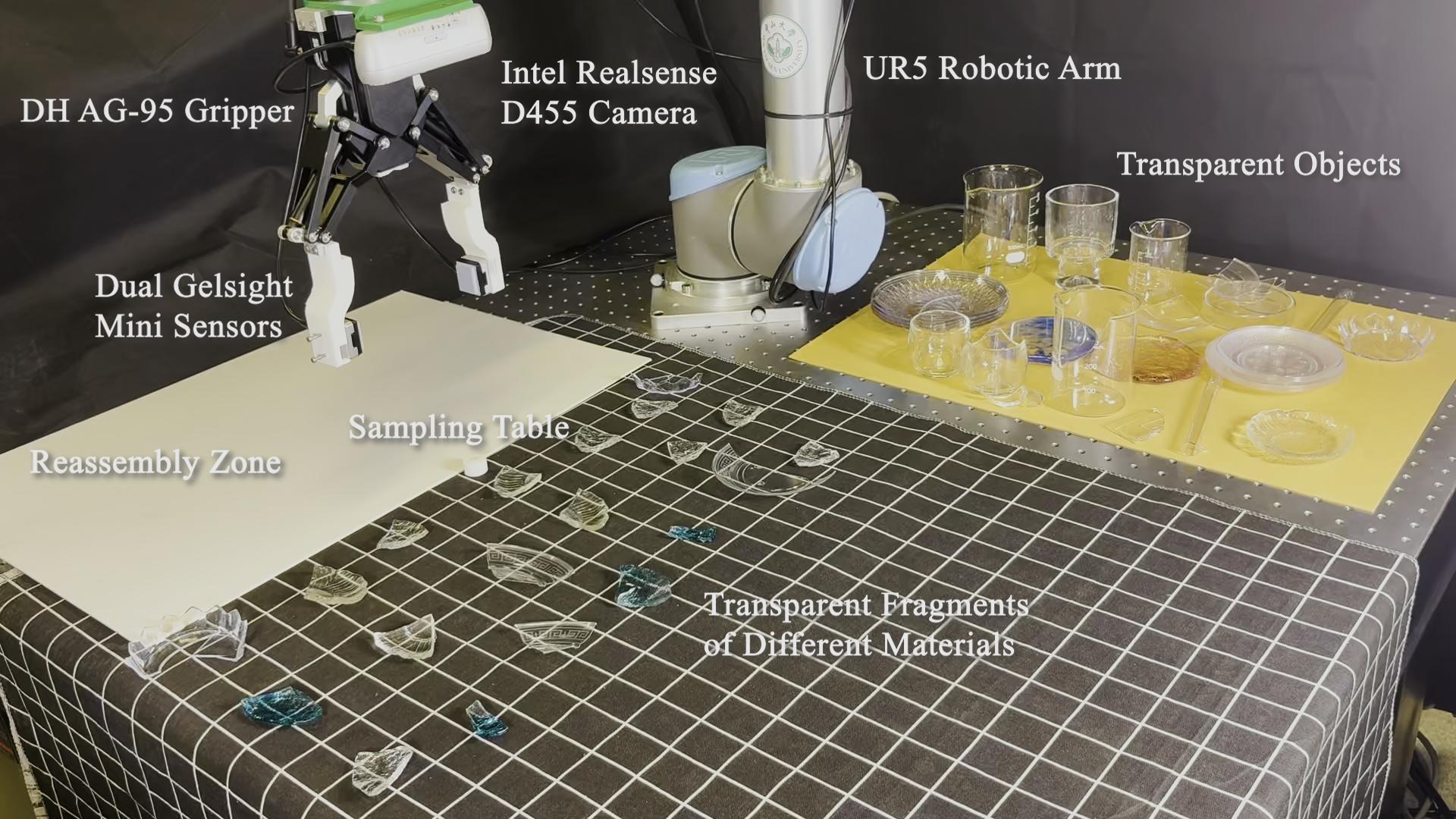}
\caption{Hardware setup: UR5 robotic arm with DH-95 two-finger gripper and integrated GelSight Mini tactile sensors.}
\label{Hardware system}
\end{figure}

Tactile sensing plays a crucial role in robotic manipulation, providing high-resolution local geometry and force feedback to complement global vision. Various tactile sensing technologies, such as piezoelectric arrays for dynamic touch sensing\cite{s16060819}, capacitive skins for large-area pressure mapping\cite{6502183}, triboelectric nanogenerators\cite{FAN2012328} for self-powered contact detection, and piezoresistive films\cite{5339133} for force measurement cannot match the spatial acuity of human fingertips. In contrast, optical tactile sensors using miniature CMOS cameras, pioneered by GelSight\cite{Dong2011GelSight}, can achieve submillimeter contact resolution at low cost. GelSight's elastomeric camera design captures detailed deformation patterns beneath an index-matched gel, allowing accurate reconstruction of surface normals and fine textures. On this basis, a series of vision-based sensors such as Geltip\cite{Gomes2020Geltip}, Digit\cite{lambeta2020Digit} and Thintact \cite{Xu2025ThinTact} have been designed. 
However, these optical methods are usually limited to discrete contact surfaces and rely on a thin layer of silicone, which is supported by a hard acrylic resin on the back, which limits their deformation range and cannot capture the complete object contour.
To address the limitation of these tactile-only methods, our system comprises a UR5 manipulator, a DH-95 two-finger gripper instrumented with dual GelSight Mini sensors, and an in-hand RGB-D camera for visual perception (as shown in Fig.~\ref{Hardware system}). Standard hand–eye and frame calibration align the RGB-D, tactile, and robot coordinates to enable synchronized data acquisition. All modules run under ROS with GPU acceleration, supporting closed-loop visual–tactile contour estimation and reassembly across diverse scenes.


\subsection{Experiment 1: Detection on Synthetic Transparent Fragments}

To validate the effectiveness of our proposed model for the task of transparent fragments contour estimation, we first conducted a rigorous performance evaluation on a custom-built synthetic dataset. This experiment was designed to assess the model's core capabilities for precise segmentation and localization of transparent objects under controlled, ideal conditions.
We selected three popular network architectures as baselines for comparison: EfficientNet \cite{tan2019efficientnet}, MobileNet \cite{Howard2019SearchingFM} and TGCNN \cite{Li2022SimTrans12K}, which has demonstrated strong performance in detection task. All models were trained and evaluated under identical experimental conditions to ensure a fair comparison. We employed the Intersection over Union (IoU) as the primary evaluation metric. This metric quantifies the overlap between the predicted mask and the ground truth, where a higher value indicates more accurate segmentation. To ensure a fair comparison, we used a unified set of training parameters for all models. The synthetic dataset used for model training consists of 6,401 training images and 1,200 testing images. A batch size of 16 was used for training. All models were trained for 100 epochs using the Adam optimizer with an initial learning rate of 1e-4 to ensure full convergence. The input image resolution was standardized to 224x224 pixels. To enhance model generalization, we applied standard data augmentation techniques during training, including random rotations, flips, and crops.

\begin{table}[h!]
\centering
\caption{Comparison of mask prediction results on the synthetic dataset and real dataset. From top to bottom: Original Image, TGCNN, EfficientNet, MobileNet, and Our Model.}
\label{tab:performance_comparison}
\begin{tabularx}{3.5in}{l >{\centering\arraybackslash}X >{\centering\arraybackslash}X >{\centering\arraybackslash}X}
\toprule
Model & Synthetic Data (IoU) & Varying Backgrounds (IoU) & Varying Illumination (IoU) \\
\midrule
EfficientNet  & 0.7875 & 0.7536 & 0.6430 \\
MobileNet     & 0.7349 & 0.6366 & 0.3205 \\
TGCNN         & 0.4576   & 0.2646 & 0.5968 \\
\textbf{Ours} & \textbf{0.8856} & \textbf{0.9382} & \textbf{0.8451} \\
\bottomrule
\end{tabularx}
\end{table}

The experimental results demonstrate that our model surpasses all baseline models in performance. As shown in the quantitative analysis in Table \ref{tab:performance_comparison}, our model achieved the highest IoU score on the test set. The qualitative results in Fig. \ref{Synthetic and Real} provide a more intuitive illustration of this. This result strongly indicates that our model can more accurately identify and segment the regions of transparent objects, providing a more reliable visual basis for subsequent grasp operations.

\subsection{Experiment 2: Detection on Real Transparent Fragments with Varying Backgrounds}

\begin{figure*}[!t]
\centering
\includegraphics[width=0.95\textwidth]{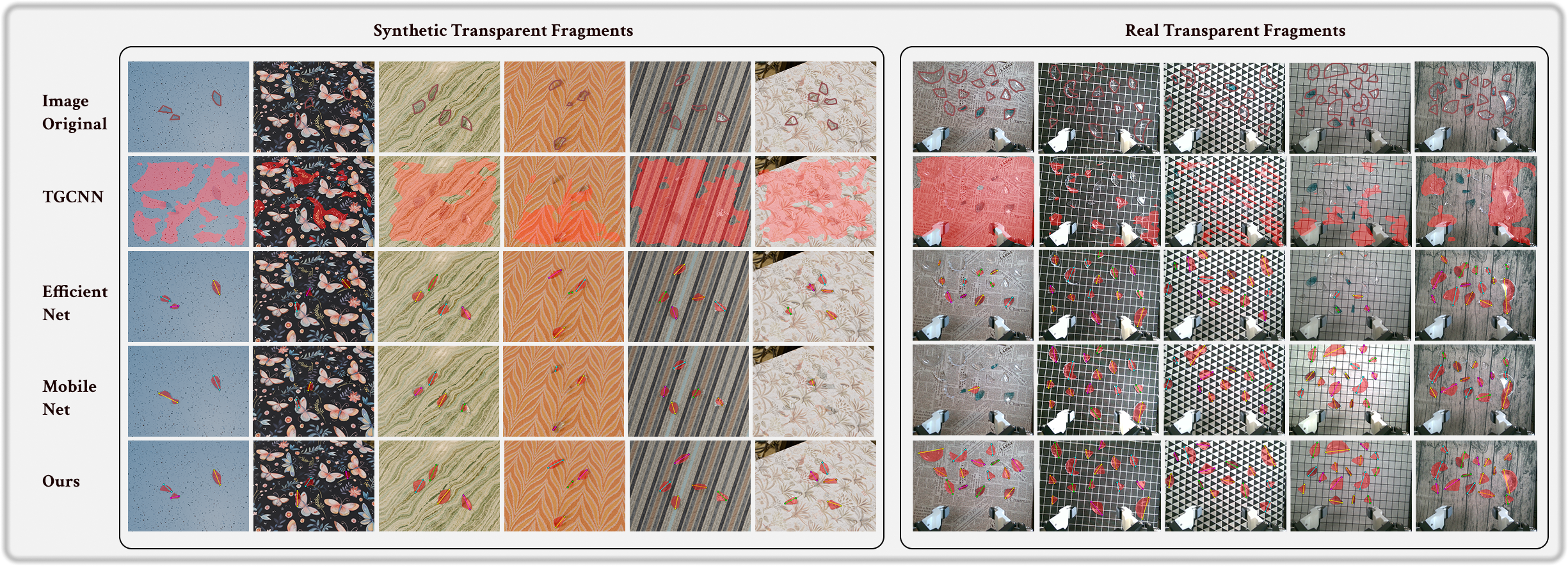}
\caption{Transparent fragment detection results and postprocessing. Including synthetic fragments detected and real fragments detected under different backgrounds.}
\label{Synthetic and Real}
\end{figure*}

\begin{figure}[!t]
\centering
\includegraphics[width=3.5in]{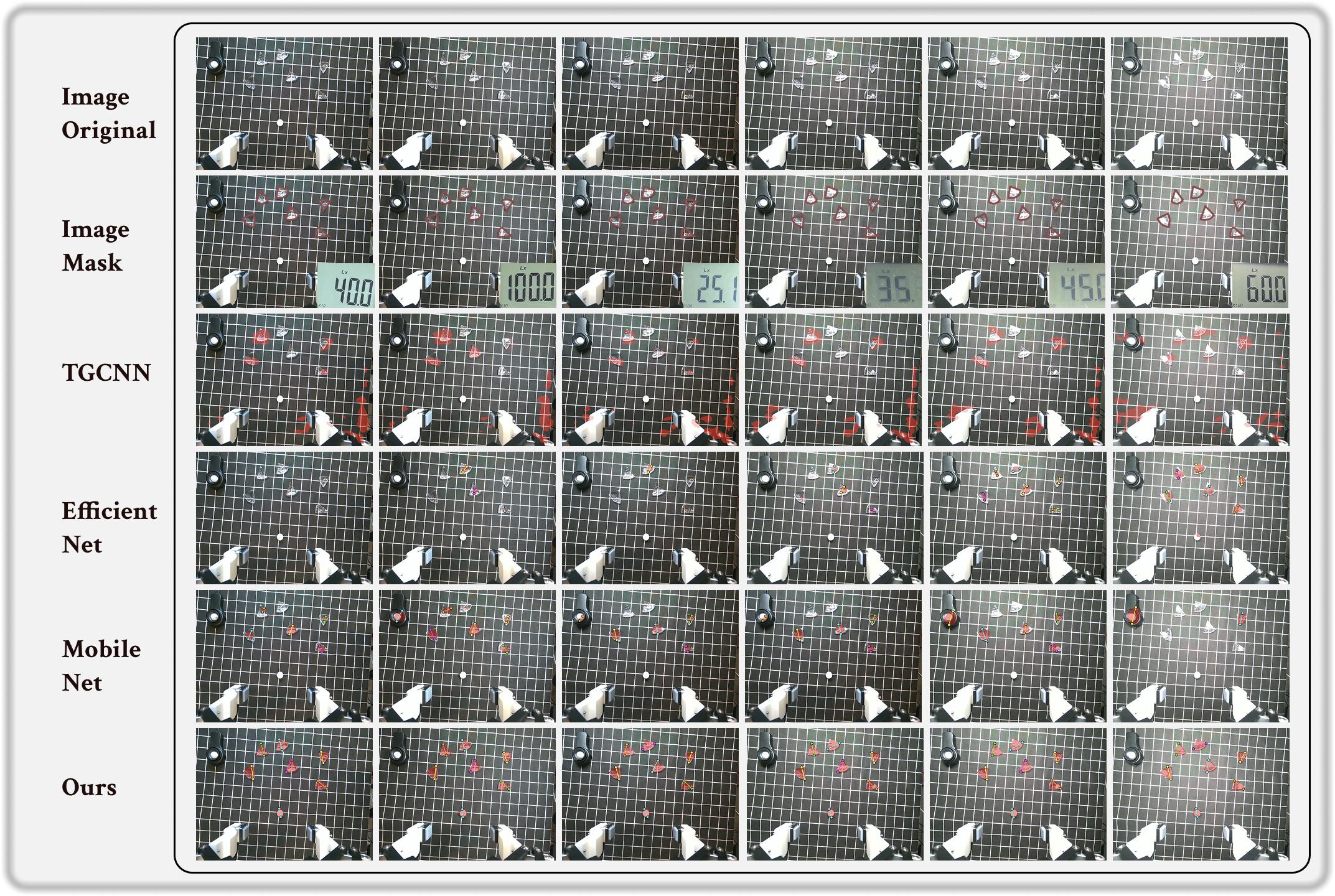}
\caption{Transparent fragment detection in varying illumination, from left to right starting from 400 to 6000 Lux.}
\label{illumination}
\end{figure}

To evaluate the robustness and generalization capabilities of our proposed detection network under real-world conditions, we select 8 backgrounds with different features, including 3 color backgrounds and 5 patterned backgrounds. We utilized a total of 15 transparent fragments made from different materials, including glass, acrylic, colored crystal, and plastic. These fragments were sequentially placed on 5 different backgrounds designed to simulate typical tabletop or industrial settings, thereby providing a comprehensive test of the model's performance. By capturing images for each fragment-background combination, we constructed a real-world test set containing 100 images. Consistent with Experiment 1, we continued to use the Intersection over Union (IoU) as the primary metric for evaluating segmentation accuracy and compared our model's performance against EfficientNet, MobileNet, and TGCNN.

As indicated by the results in Table \ref{tab:performance_comparison}, our model once again demonstrated superior performance on the real-world test data. The qualitative results in Fig. \ref{Synthetic and Real} provide a more intuitive illustration of this. When faced with complex backgrounds or specular reflections from the fragments, the performance of EfficientNet and TGCNN degraded substantially, leading to numerous incorrect segmentation areas. MobilNet performed relatively better but still exhibited incomplete masks when processing highly reflective glass or acrylic fragments with indistinct edges. 
In contrast, our model consistently handled the challenges posed by different materials, such as strong reflections from glass and color interference from colored crystal, and complex background textures, generating complete and precise segmentation masks. These results affirm that our model not only excels on synthetic data but also possesses strong generalization ability and stability in challenging physical-world scenarios.

\subsection{Experiment 3: Detection on Real Transparent Fragments with Varying Illumination}

To further examine the stability and robustness of our model in complex lighting environments, we designed a third set of experiments. Variations in lighting conditions are a critical factor affecting the performance of vision algorithms in real-world deployments, as they can significantly alter image contrast, introduce noise and reflections, and thus pose a severe challenge to the accuracy of transparent fragment detection networks. In this experiment, we selected 6 typical transparent fragments and placed them on a fixed background. By precisely controlling an external light source, 
we captured images under six different illumination intensities including 400, 1000, 2500, 3500, 4500, and 6000 Lux, to construct a dataset specifically for illumination robustness testing. This range of intensities covers various common industrial and daily-life scenarios, from dim to bright.

The experimental results in Table \ref{tab:performance_comparison} clearly reveal the varying sensitivity of the models to changes in illumination. As shown in Table I, under the challenging conditions of low light (400 Lux) and high light (6000 Lux), the performance of the baseline models degraded significantly. Low illumination reduced the signal-to-noise ratio of the images, making it difficult for EfficientNet and MobileNet to effectively distinguish the object contours. Conversely, high illumination caused overexposure and strong specular reflections on the surfaces of the transparent fragments, severely interfering with the detection capabilities of all baseline models. In contrast, our model maintained a consistently high IoU score across the entire range of tested illumination levels, with far less performance fluctuation than the other models. The qualitative results in Fig. \ref{illumination} also visually corroborate this, showing that our model continues to generate accurate and complete segmentation masks in both dim and bright environments.

In summary, the results of Experiment 3 provide strong evidence of our proposed model's excellent robustness to variations in illumination. Its ability to maintain stable and reliable detection performance across a wide dynamic range of lighting conditions is of critical importance for its deployment in uncontrolled, real-world environments.

\subsection{Experiment 4: Partial Fracture Reassembly via visual contour-Based Matching}

\begin{figure}[!t]
\centering
\includegraphics[width=3.4in]{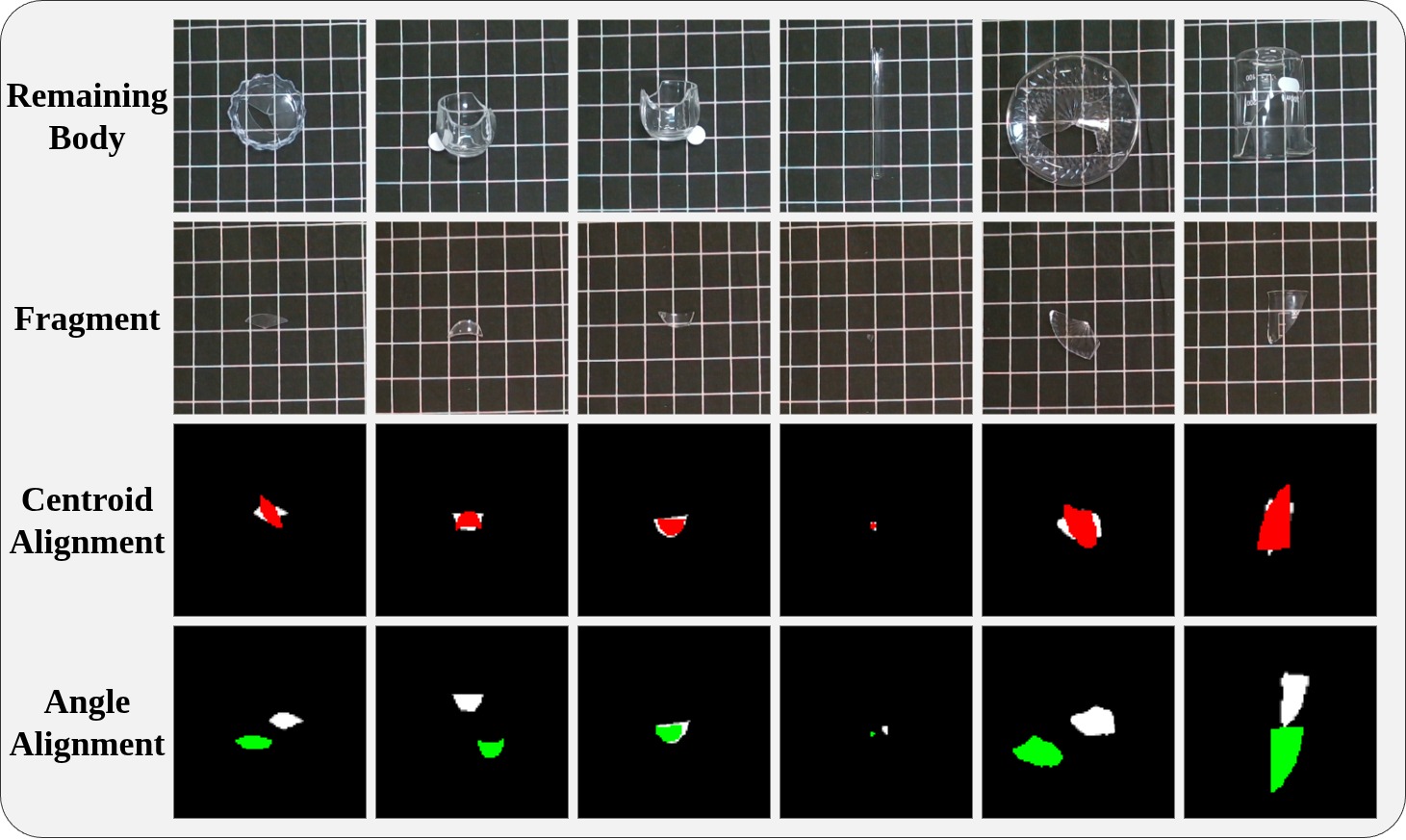}
\caption{Visual shape-based partial fracture reassembly results: five representative cases with different notched transparent objects, showing remaining body, detached fragment, centroid alignment, and final angle-aligned overlay.}
\label{partial}
\end{figure}

To validate the proposed visual-only contour matching pipeline for partially fractured transparent objects, we designed representative test cases where an object's main body remains intact while one or two fragments are detached.
To test a range of real-world fracture scenarios, we selected diverse transparent objects with partial damage. As shown in the Fig. \ref{partial}, from left to right: an acrylic dish with a notch in the middle, two notches in the same glass, a test tube with a slight notch at the mouth, an acrylic plate with a notch in the middle, and a notched beaker, each case includes the remaining body, the detached fragment, and visualizations of our mask alignment process. Using our trained detection network TransFragNet and the Segment Anything Model (SAM), we extract masks of both the fragment and the corresponding gap on the main body. The pipeline then applies thresholding, resizing, and centroid alignment (visualized in red) to normalize shapes for comparison.
Next, a rotation sweep is performed to find the optimal orientation. The final alignment is confirmed by comparing centroid overlap and angle-adjusted shape alignment (green/white overlay in the figure). At each rotation step, the Intersection over Union (IoU) and Chamfer distance are computed, and their weighted combination provides the final matching score.

\begin{table}[h!]
\centering
\caption{Results for Partial Fracture Reassembly}
\label{tab:exp4}
\begin{tabularx}{3.5in}{l >{\centering\arraybackslash}X >{\centering\arraybackslash}X}
\toprule
Metric & IoU & Chamfer Distance \\
\midrule
Pair 1 & 0.82 & 4.3 px \\
Pair 2 & 0.79 & 5.1 px \\
Pair 3 & 0.85 & 3.8 px \\
Pair 4 & 0.81 & 2.4 px \\
Pair 5 & 0.74 & 4.5 px \\
Pair 6 & 0.82 & 5.2 px \\
\bottomrule
\end{tabularx}
\end{table}

Table \ref{tab:exp4} reports the IoU and Chamfer distance for six representative fragment-body pairs. Together, these results visually and quantitatively demonstrate that our vision-based approach achieves accurate alignment for partially fractured transparent objects, even without tactile input. This is especially practical when the object is too large or fragile to be fully grasped and sampled.
This also provides preliminary validation for our database matching algorithm based on visual indices. This experiment demonstrates the possibility of reliable shape estimation based solely on visual contour in scenarios where precise tactile perception is impractical, such as for large, nearly complete transparent objects.

\subsection{Experiment 5: Complete Fracture Reassembly with Visual-Tactile Fusion}

\begin{figure}[!t]
\centering
\includegraphics[width=3.4in]{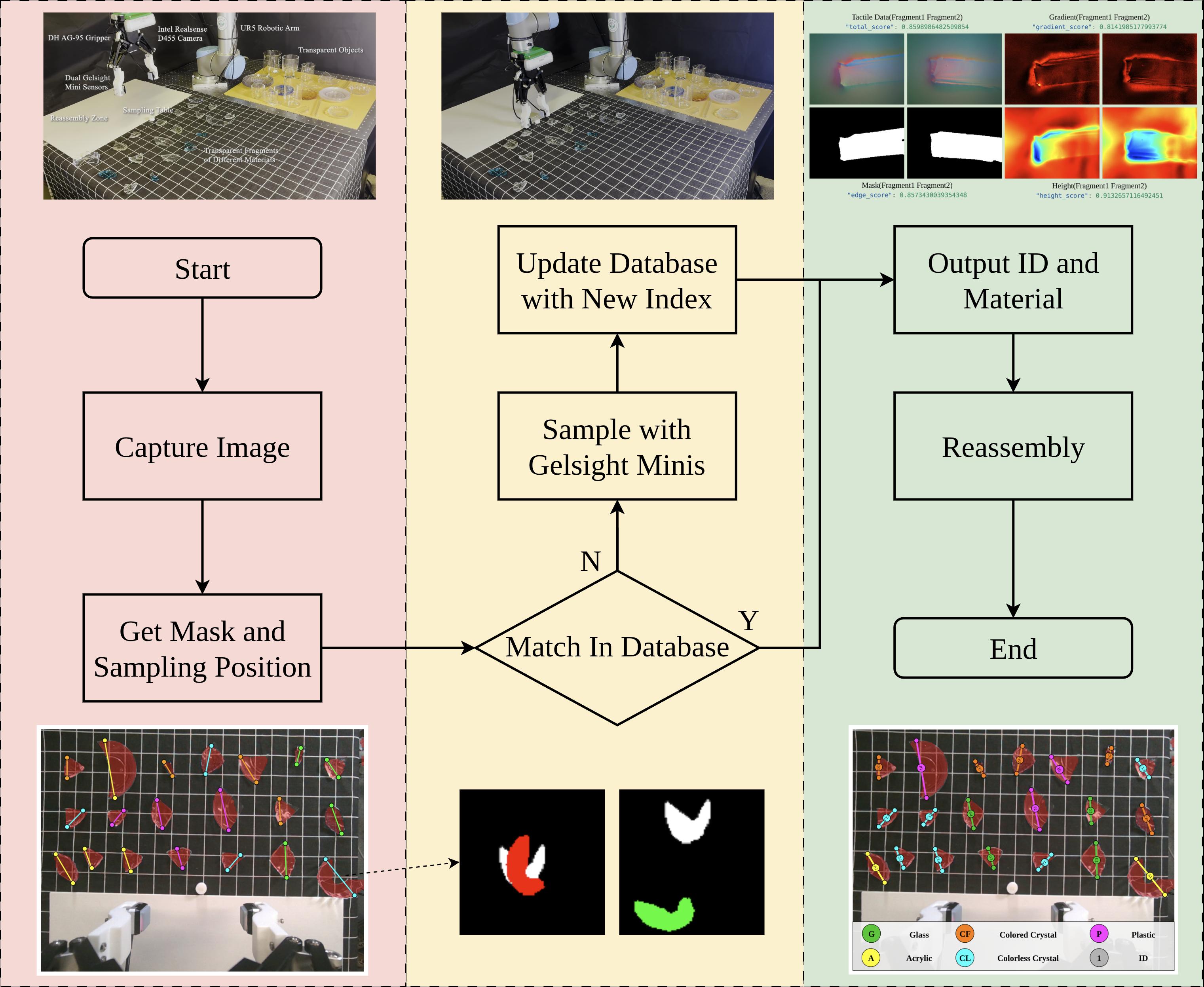}
\caption{Workflow of the proposed visual–tactile transparent fragment reassembly framework. The framework begins with visual detection using an RGB-D camera to segment transparent fragments and estimate grasping positions using TransFragNet. Fragments' extracted visual features are then matched in the database. If a match is not found in the existing database, new tactile features sampling process including edge maps, gradient fields, and height profiles are extracted and stored. Matched fragments are classified by material and assigned IDs based on multi-dimensional feature similarity, then passed to the reassembly stage.}
\label{fig:flowchart}
\end{figure}

\begin{figure*}[!t]
\centering
\includegraphics[width=0.9\textwidth]{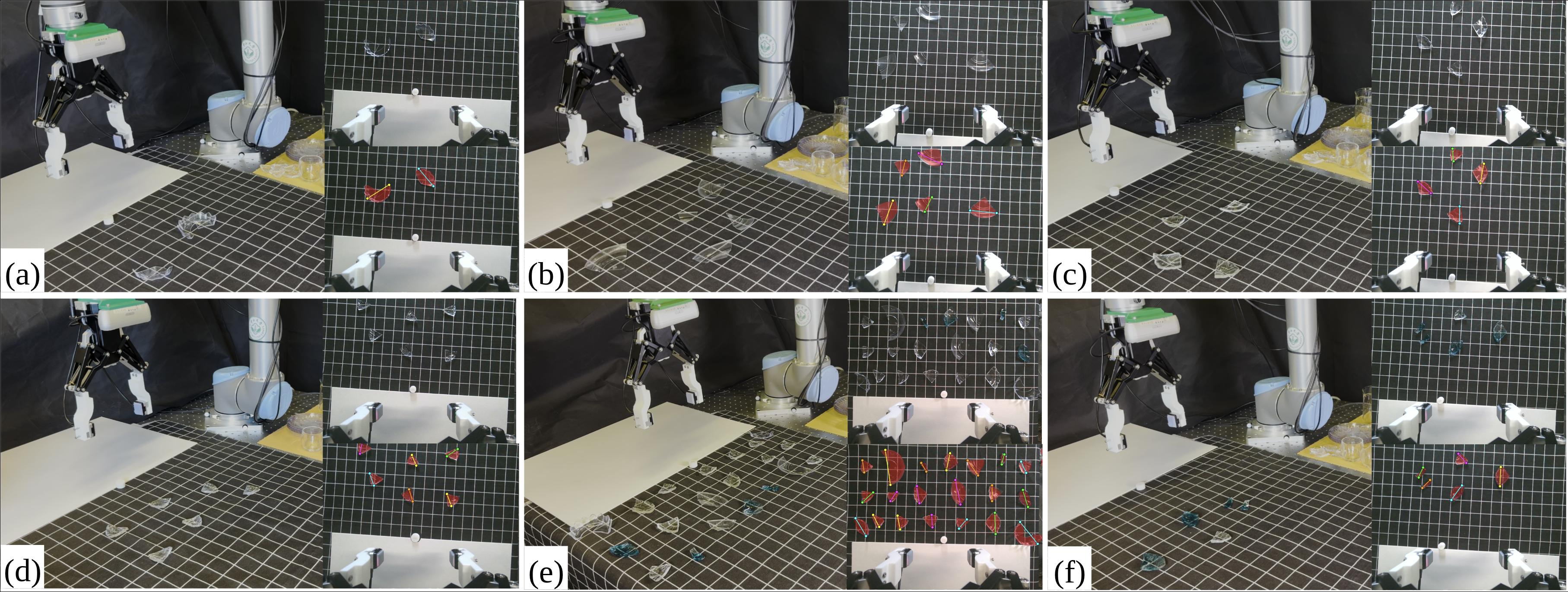}
\caption{Scenes of representative fragments for complete fracture reassembly: (a) acrylic dish; (b) plastic dish; (c) glass dish; (d) colorless crystal dish; (e) mixed-material dish fragments; (f) colored crystal dish.}
\label{single0}
\end{figure*}

To verify the feasibility of our contour estimation framework based reassembly strategy, we designed experiments covering 5 transparent materials and 25 complete Fracture fragments, under random placement situation. Fig. \ref{single0} illustrates typical scenes from acrylic, plastic, glass, colorless crystal, colored crystal, and mixed-material dish fragments.

\begin{table}[h!]
\centering
\caption{Visual-Tactile Fusion Reassembly Scores for Different Materials}
\label{tab:exp5_fusion}
\begin{tabularx}{3.5in}{l >{\centering\arraybackslash}X >{\centering\arraybackslash}X >{\centering\arraybackslash}X >{\centering\arraybackslash}X}
\toprule
Material & Edge Score & Gradient Score & Height Score & Total Fusion \\
\midrule
Glass (Pair 1) & 0.857 & 0.814 & 0.913 & 0.859 \\
Glass (Pair 2) & 0.887 & 0.908 & 0.891 & 0.892 \\
Acrylic (Pair 1) & 0.844 & 0.880 & 0.917 & 0.865 \\
Colored Crystal (Pair 1) & 0.836 & 0.932 & 0.818 & 0.852 \\
Plastic (Pair 1) & 0.846 & 0.870 & 0.936 & 0.869 \\
Colorless Crystal (Pair 1) & 0.887 & 0.884 & 0.906 & 0.890 \\
\bottomrule
\end{tabularx}
\end{table}

\begin{figure*}[!t]
\centering
\includegraphics[width=0.9\textwidth]{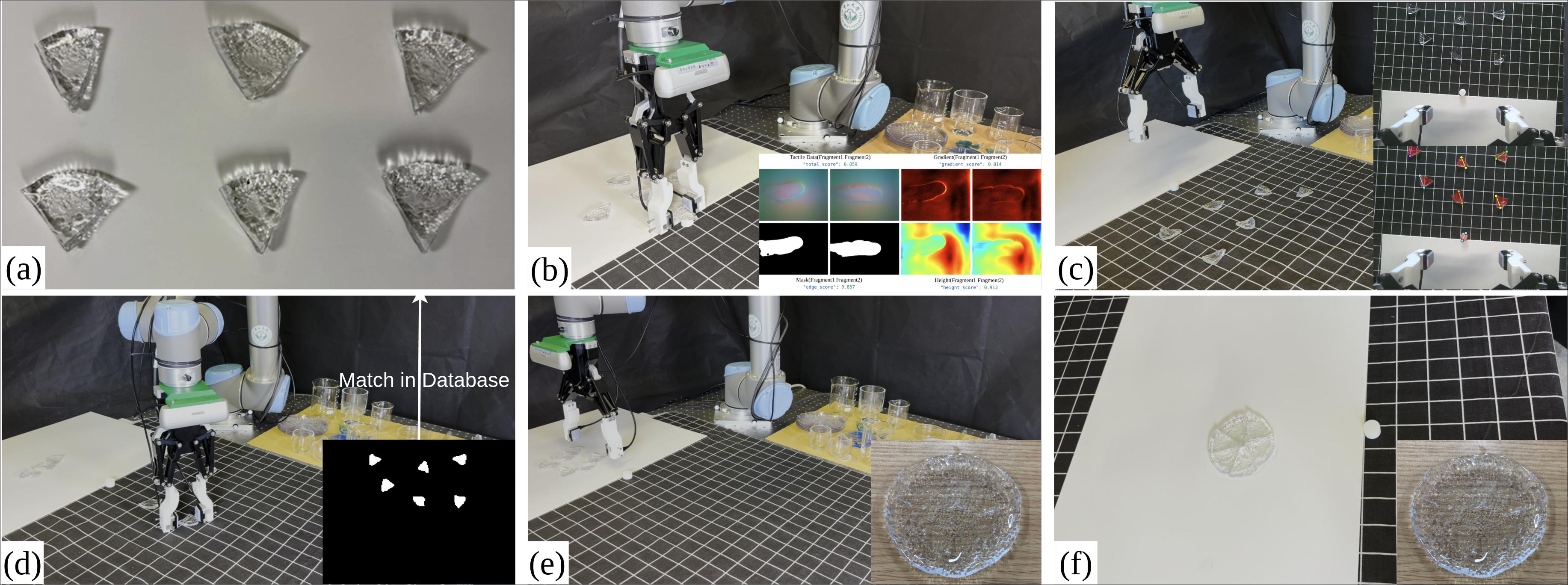}
\caption{Experimental pipeline for complete reassembly with visual-tactile fusion, demonstrated with a colorless crystal dish: (a) full fragment set; (b) edge sampling and reconstruction; (c) random placement scene; (d) match visual ID with Database and excute reassembly; (e-f) final reassembled result with different perspectives.}
\label{single1}
\end{figure*}

As shown in Fig. \ref{single1}, we take the colorless crystal dish as an example to describe the whole pipeline. First, the complete set of fragments is imaged and visually segmented; the fragment masks are extracted to create an initial index in the database. Next, each fragment's fracture edge is sampled with the GelSight Mini sensors to acquire raw tactile data, which are then reconstructed into edge masks, local height maps, and gradient fields. The tactile information is fused with the visual mask for material classification and improved matching. Using the contour matching strategy, the fragments are indexed and the reassembly sequence is determined. Finally, the fragments are placed randomly on the table, visually detected, identified by index, and matched in order. Fig. \ref{single1} (e-f) show the physical reassembly results for the selected object.

The quantitative results of contour matching in Section \ref{Subsection: IV.D}, Fig. \ref{fig:edgematch} are summarized in Table \ref{tab:exp5_fusion}, demonstrating the consistent matching accuracy across different materials and confirming that the integrated contour estimation framework can handle diverse complete fracture cases. The final matching scores demonstrate the complementary benefit of fusing edge, gradient, and height cues, enabling robust contour estimation even for geometrically ambiguous transparent fragments.

\subsection{Experiment 6: Complete Fracture Reassembly on Mixed-Material Transparent Fragments Scene}

\begin{figure*}[!t]
\centering
\includegraphics[width=0.9\textwidth]{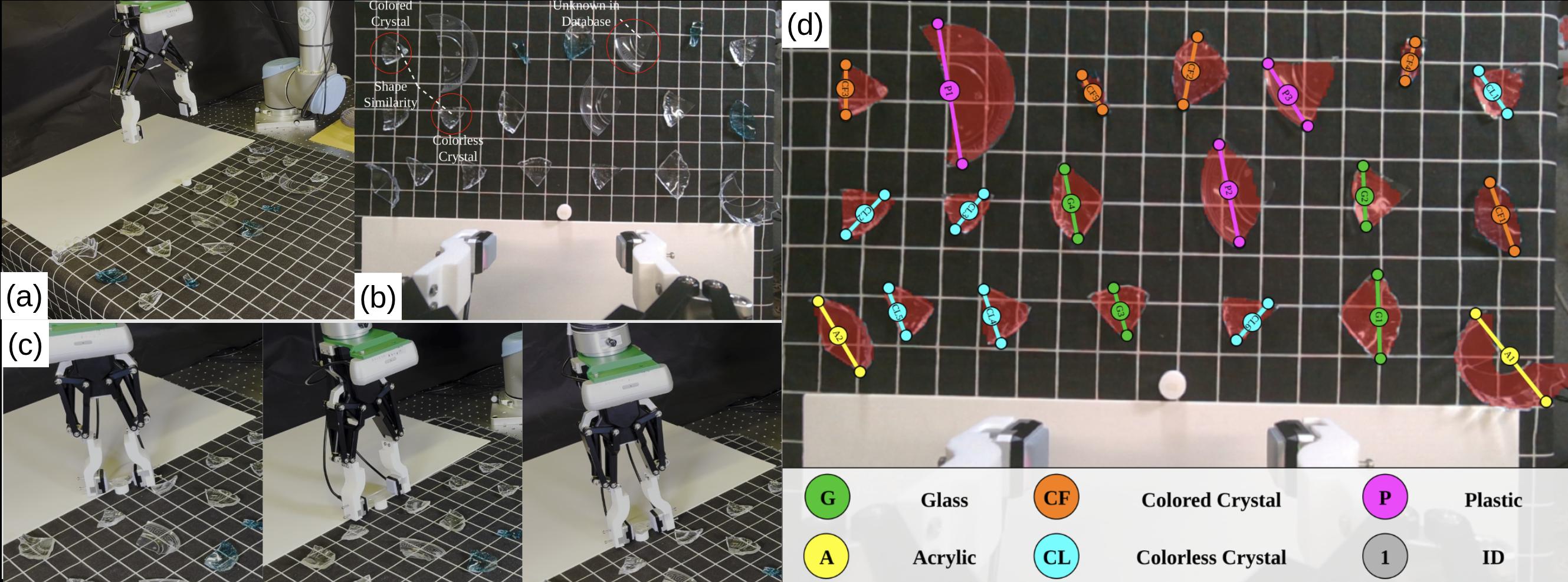}
\caption{Generalization test for complete fracture reassembly with unseen or similar transparent fragments. (a) Experimental scene with mixed-material fragments including glass, acrylic, and colored/colorless crystal pieces. (b) Top view with highlighted examples: CF3 (colored crystal) and CL3 (colorless crystal) as shape-similar pairs; P3 as an unknown plastic fragment not stored in the database. (c) Edge sampling procedure using the GelSight Mini sensor for fallback classification. (d) Final identification map showing each fragment's assigned material class and unique ID, providing full guidance for robust reassembly.}
\label{mixed0}
\end{figure*}

\begin{figure*}[!t]
\centering
\includegraphics[width=0.9\textwidth]{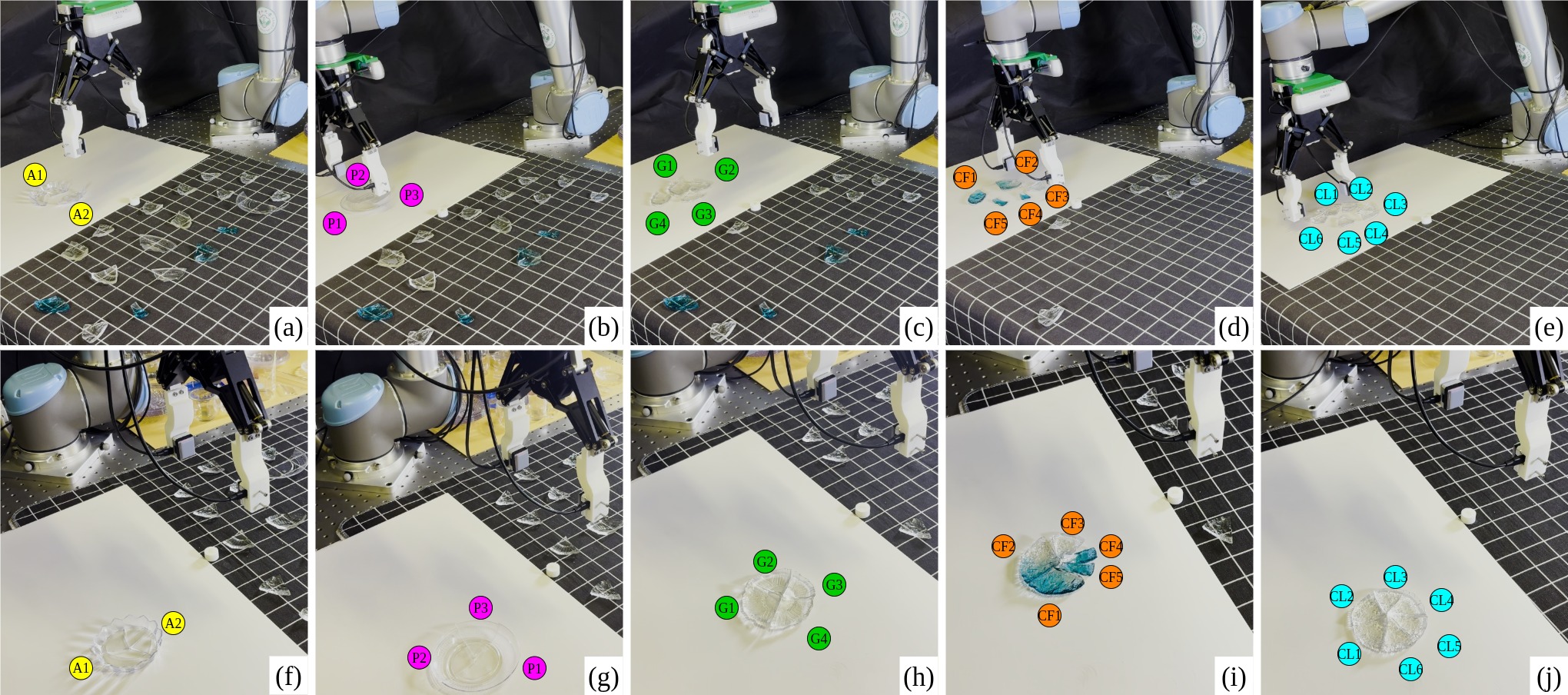}
\caption{Reassembly results of mixed fragments from completely fractured transparent objects made from different transparent materials: (a) acrylic dish; (b) plastic dish; (c) glass dish; (d) colorless crystal dish; (e) colored crystal dish. (f-j) the same as (a-e) but with different perspectives.}
\label{mixed1}
\end{figure*}

To test the method's generalization ability, we further design mixed fragment scenes of 20 fragments from 5 different materials containing pieces that are not found in the database or cause ambiguous matches due to highly similar shapes. In these cases, the fallback strategy is triggered: the unmatched or ambiguous fragment is resampled on-site with Gelsight Mini sensors to capture raw contact data. The new tactile signal is reconstructed into edge, gradient, and height features, and the same contour matching pipeline is used to find the best alignment with other fragments. Additionally, material classification is performed on the sampled fragment to restrict matching to candidates of the same material type, improving discrimination for visually similar pieces. This fallback mechanism ensures that robust reassembly is achievable even for novel or unseen fractures and demonstrates how the database can be incrementally expanded with new tactile profiles to support future tasks.

\begin{table}[h!]
\centering
\caption{Performance in Mixed Scene with Similar and Unseen Fragments}
\label{tab:exp6}
\begin{tabularx}{3.5in}{l >{\centering\arraybackslash}X >{\centering\arraybackslash}X >{\centering\arraybackslash}X >{\centering\arraybackslash}X}
\toprule
Scene Description & Top-1 Acc. & Top-3 Acc. & Mean IoU & Fallback Needed \\
\midrule
Similar-Look Case (CL3 vs CF3) & 50\% & 100\% & 0.79 & Yes \\
Unknown Fragment (P3) & 0\% & --- & --- & Yes \\
Full Mixed Material Scene & 85\% & 100\% & 0.81 & Resolved \\
\bottomrule
\end{tabularx}
\end{table}

Fig. \ref{mixed0} illustrates the challenging mixed-fragment scenario evaluated in Experiment 6. When visually similar or previously unseen fragments such as Colored Crystal Fragment 3 (CF3), Colorless Crystal Fragment 3 (CL3), and Plastic Fragment 3 (P3) could not be resolved through the initial visual-tactile index match, the fallback sampling stage was triggered. The edge profiles were re-sampled with the GelSight Mini sensor, and material classification plus shape refinement re-identified the correct classes and IDs.
As shown in Table~\ref{tab:exp6}, for the full mixed material scene, the Top-1 retrieval accuracy reached 85\%, and Top-3 accuracy reached 100\%. The few ambiguous cases involved fragments with nearly identical shape and material, which were successfully resolved by our fallback tactile sampling and reclassification step. This demonstrates that the proposed visual-database matching generalizes well even in dense mixed-fragment scenes and supports reliable reassembly when integrated with tactile fusion.
After the fallback strategy, we performed the reassembly experiment in the same way as Experiment 5. The final experimental results are shown in Fig. \ref{mixed1}. The transparent fragments of the five materials were successfully reassembled.

\subsection{Limitations}
While our framework demonstrates robust performance across flat and some curved transparent fragments, several limitations remain. First, the performance may degrade when the fragment has high curvature or complex intersections, such as fully fractured goblets, beakers, or fragments with sharp corner regions. These geometries can cause local ambiguities in mask extraction and tactile sampling, especially at tight curves and thin edges. This highlights an important next step for expanding our benchmark to cover more diverse fragment shapes and refining our contour estimation and matching algorithms to better handle highly curved transparent geometries. Failure cases were observed when multiple fragments of similar size and contour appear in a cluttered scene. In these instances, the framework may misclassify or mismatch fragments with very subtle edge differences, which can propagate errors in the reassembly order. To address this, fallback conditions and additional tactile refinement have been integrated, but edge cases still pose challenges in extremely complex break scenarios. In addition, complex real-world situation will affect the stability of the framework. Future work will focus on these aspects, aiming to expand the benchmark and dataset to include more use cases and improve the robustness of the full framwork in practical applications.

\section{Conclusion}

To address the challenging problem of contour estimation and reassembling transparent irregular fragments, this paper proposes a visual–tactile fusion benchmark framework that integrates synthetic and real-world data, visual detection, high-resolution tactile sensing, contour matching and reassembly.
First, we constructed TransFrag27K, a large-scale transparent fragment dataset including diverse synthetic break scenarios. By leveraging the Blender simulation pipeline and refined annotation strategy, the dataset provides high-quality masks and robust training data for detection under various backgrounds and illumination conditions.
Next, we proposed TransFragNet, a dedicated grasping position detection network that achieves accurate mask segmentation and sampling prediction for transparent fragments. Experiments show that TransFragNet maintains strong detection performance across synthetic and real scenes, even under challenging lighting or background.
To compensate for the limited geometric information in visual signals, we employed a two-finger gripper equipped with GelSight Mini tactile sensors to reconstruct fine edge and height features along fragment surfaces. 
By fusing this tactile information with visual cues, our framework achieves reliable material classification and final contour estimation, which directly supports precise fragment matching and autonomous reassembly, and demonstrates strong performance in real-world validation.
Our experiments validate the effectiveness of the proposed framework, covering scenarios from partial fracture reassembly using vision-only alignment, to complete fracture tasks requiring visual–tactile fusion, and robustness tests on mixed-material fragment scenes.

Overall, this work establishes a general benchmark for transparent fragment detection, contour estimation, matching and reassembly. It demonstrates the potential of visual–tactile fusion in handling challenging transparent fragments where conventional methods fail. Future work will expand the dataset with more complex fracture types and develop adaptive tactile exploration strategies to further enhance real-world generalization.

\bibliographystyle{IEEEtran}
\bibliography{references}


 



\vspace{-5.9em} 
\begin{IEEEbiography}[{\includegraphics[width=1in,height=1.25in,clip,keepaspectratio]{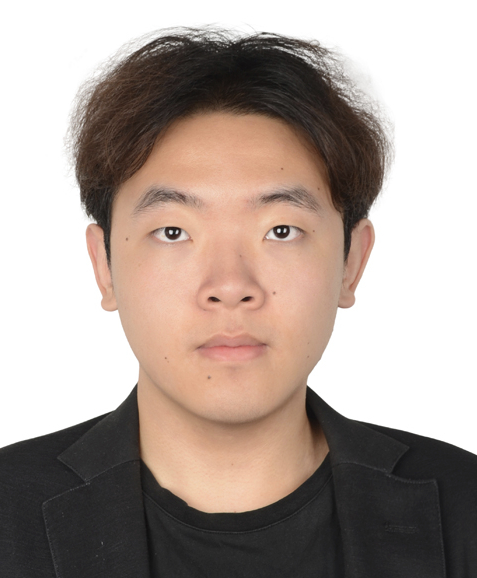}}]{Qihao Lin}
received the B.Eng degree from School of Advanced Manufacturing, Sun Yat-sen University in 2025. He is an assistant researcher in School of Advanced Manufacturing, Sun Yat-sen University. His research interests include robotics, robotic vision and manipulation, machine learning.\end{IEEEbiography}
\vspace{-2.9em} 

\begin{IEEEbiography}[{\includegraphics[width=1in,height=1.25in,clip,keepaspectratio]{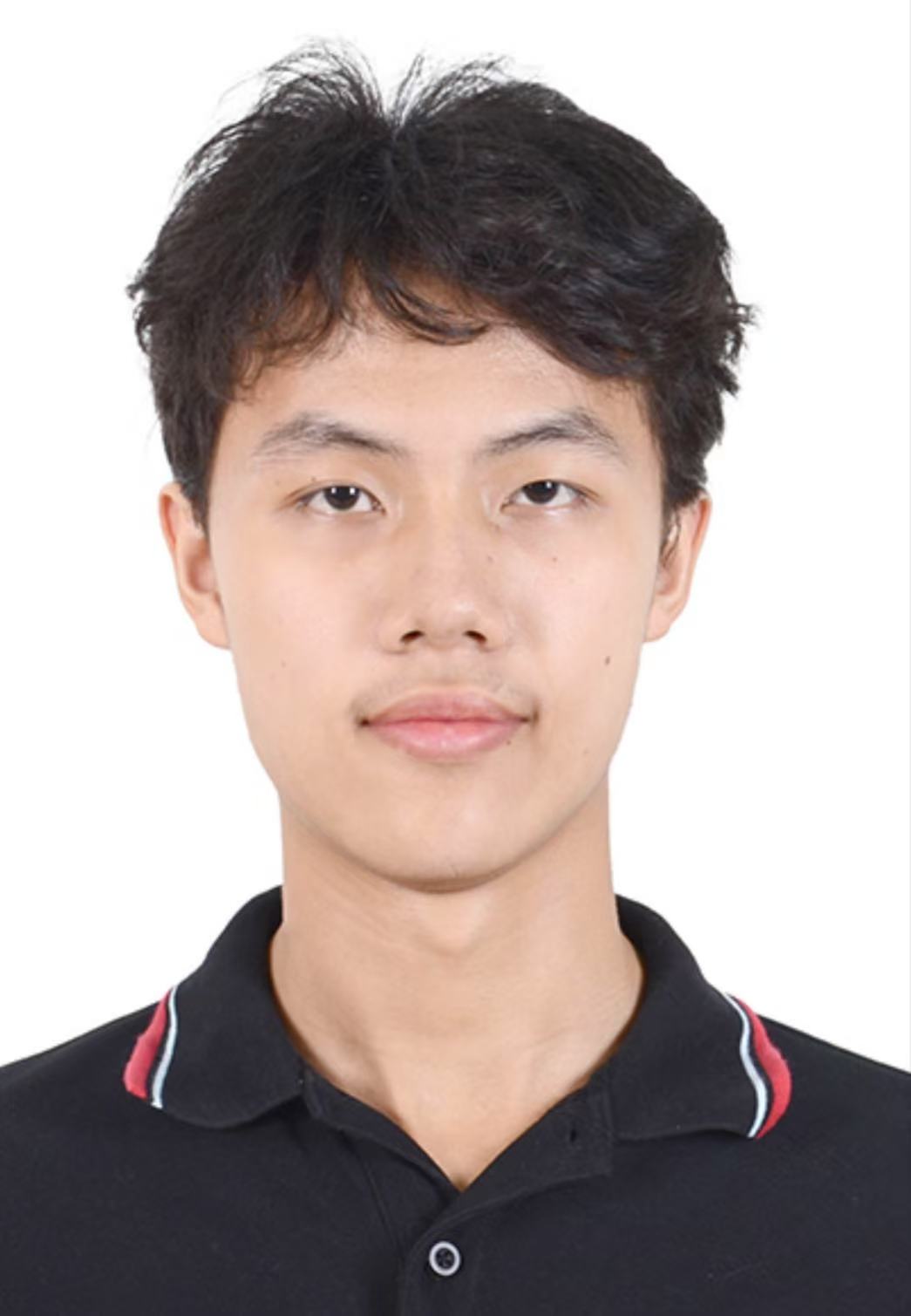}}]{Borui Chen}
is currently working toward the B.Eng. degree in mechanical engineering with the School of Advanced Manufacturing, Sun Yat-sen University. His research interests include multi-modal perception, machine learning and robotics.\end{IEEEbiography}
\vspace{-2.9em}

\begin{IEEEbiography}[{\includegraphics[width=1in,height=1.25in,clip,keepaspectratio]{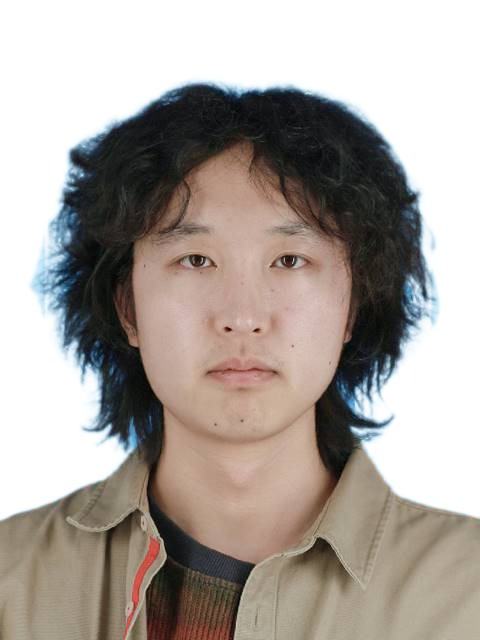}}]{Yuping Zhou}
is an undergraduate student at the School of Intelligent Systems Engineering, Sun Yat-sen University, and is currently a visiting student at The Hong Kong University of Science and Technology (Guangzhou). His research interests include computer vision and embodied intelligence.\end{IEEEbiography}
\vspace{-2.9em} 

\begin{IEEEbiography}[{\includegraphics[width=1in,height=1.25in,clip,keepaspectratio]{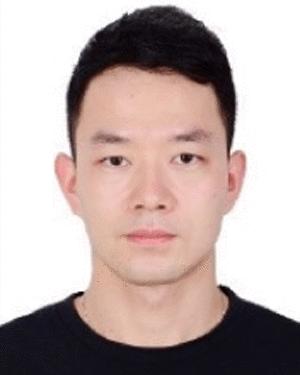}}]{Jianing Wu}
received the Ph.D. degree in mechanical engineering from the Tsinghua University, Beijing, China, in 2015.
He is currently an Associate Professor with the School of Advanced Manufacturing, Sun Yat-Sen University, Shenzhen, China. He has authored or coauthored more than 50 publications in journals, book chapters, and conference proceedings. His research interests include mechanical system reliability, biological intelligence mechanism, and bioinspired robotics.\end{IEEEbiography}
\vspace{-2.9em} 

\begin{IEEEbiography}[{\includegraphics[width=1in,height=1.25in,clip,keepaspectratio]{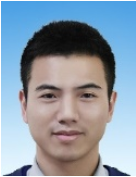}}]{Yulan Guo}
is a Senior Member of IEEE, and received the BE and PhD degrees from the National University of Defense Technology (NUDT), in 2008 and 2015, respectively. He has authored more than 200 articles with highly referred journals and conferences. His current research interests focus on 3D vision, particularly on 3D feature learning, 3D modeling, 3D object recognition, and scene understanding. He served as an associate editor for IEEE Transactions on Image Processing, the Visual Computer, and Computers \& Graphics. He also served as an area chair for CVPR, ICCV, ECCV, NeurIPS, and ACM Multimedia. He is a senior member of ACM.\end{IEEEbiography}
\vspace{-2.9em} 

\begin{IEEEbiography}[{\includegraphics[width=1in,height=1.25in,clip,keepaspectratio]{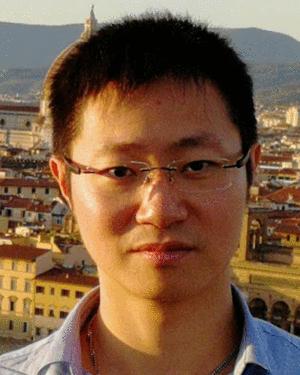}}]{Weishi Zheng}
is now a Full Professor with Sun Yat-sen University. He has now published more than 200 papers, including more than 150 publications in main journals (TPAMI, IJCV, SIGGRAPH, TIP) and top conferences (ICCV, CVPR, ECCV, NeurIPS). His research interests include person association and activity understanding, AI robotics learning, and the related weakly supervised/unsupervised and continuous learning machine learning algorithms. He has ever served as area chairs of ICCV, CVPR, ECCV, BMVC, and NeurIPS. He is an Associate Editors/on the editorial board of Artificial Intelligence Journal, Pattern Recognition. He has ever joined Microsoft Research Asia Young Faculty Visiting Programme. He is a Cheung Kong Scholar Distinguished Professor, a recipient of the NSFC Excellent Young Scientists Fund, a Fellow of IAPR, and a recipient of the Royal Society-Newton Advanced Fellowship of U.K.\end{IEEEbiography}
\vspace{-2.9em} 

\begin{IEEEbiography}[{\includegraphics[width=1in,height=1.25in,clip,keepaspectratio]{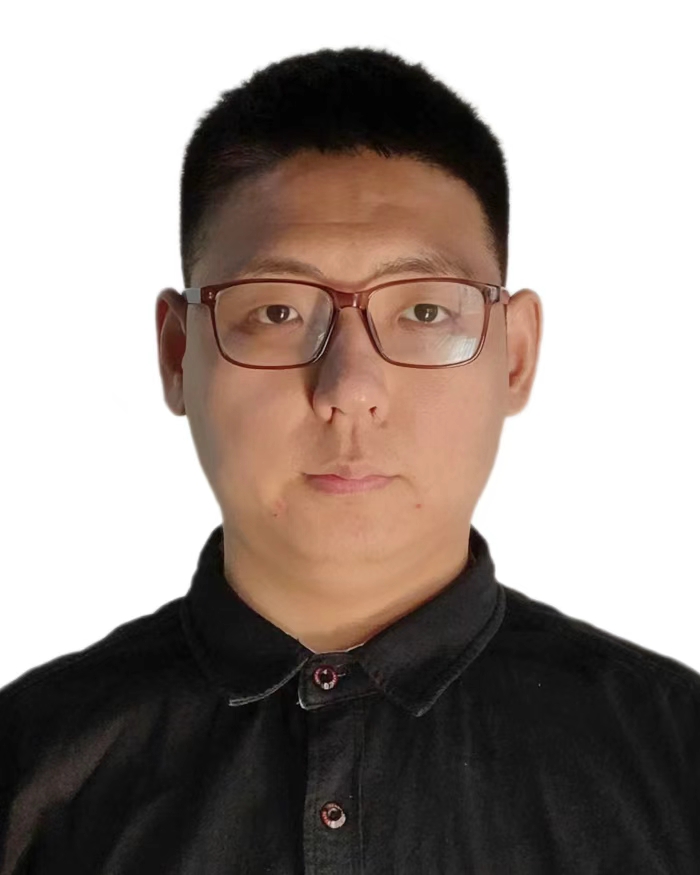}}]{Chongkun Xia}
received the Ph.D. degree from Northeastern University, Shenyang, China. Now he is an associate professor with School of Advanced Manufacturing, Sun Yat-sen University. And he also is a leader of Robot Technology \& Intelligent Perception Lab.
He has authored or coauthored more than 50 publications in SCI/EI academic papers in the field of robotics and computer vision. His research interests include robotics, robot vision and manipulation, machine learning.\end{IEEEbiography}

\vfill

\end{document}